\pdfoutput=1
\documentclass[11pt]{article}

\usepackage[final]{acl}

\usepackage{times}
\usepackage{latexsym}

\usepackage[T1]{fontenc}
\usepackage[utf8]{inputenc}

\usepackage{microtype}

\usepackage{inconsolata}

\usepackage{longtable}
\usepackage{xcolor}
\usepackage{colortbl}
\usepackage{multicol}
\usepackage{booktabs}
\usepackage{makecell}
\usepackage{amsmath, amssymb}
\usepackage{multirow}
\usepackage{mathtools}
\usepackage[inline]{enumitem}
\usepackage{subcaption}
\usepackage{float}
\usepackage{graphicx}
\usepackage{caption} 
\usepackage{upgreek}
\usepackage{seqsplit}
\usepackage{color, soul}
\usepackage{arydshln}
\usepackage{xspace}
\usepackage{comment}
\usepackage{url}
\usepackage{relsize}
\usepackage[most]{tcolorbox}
\tcbuselibrary{breakable}

\usepackage{amsmath,amsfonts,bm}

\def\eqref#1{equation~\ref{#1}}
\def\1{\bm{1}}

\def\va{{\bm{a}}}

\def\vd{{\bm{d}}}

\def\vq{{\bm{q}}}

\def\vs{{\bm{s}}}
\def\vt{{\bm{t}}}

\DeclareMathAlphabet{\mathsfit}{\encodingdefault}{\sfdefault}{m}{sl}
\SetMathAlphabet{\mathsfit}{bold}{\encodingdefault}{\sfdefault}{bx}{n}

\usepackage{caption}
\usepackage{subcaption}
\newcolumntype{Y}{>{\centering\arraybackslash}m{0pt}} 
\newcommand{\ours}{\textsc{ToTAL}\xspace}

\usepackage{xspace}
\usepackage{tcolorbox}
\usepackage[normalem]{ulem}

\title{When Thoughts Meet Facts: Reusable Reasoning for Long-Context LMs}

\author{
    Soyeong Jeong$^1$\thanks{Work done during internship at Amazon.}
    \; Taehee Jung$^2$
    \; Sung Ju Hwang$^1$
    \; Joo-Kyung Kim$^2$
    \; Dongyeop Kang$^3$
    \\
    KAIST$^{1}$ \;\; Amazon$^{2}$ \;\; University of Minnesota$^{3}$\\
   \texttt{\{starsuzi, sungju.hwang\}@kaist.ac.kr,}\\
   \texttt{\{{jungtaeh, jookyk}\}@amazon.com, {dongyeop}@umn.edu}
}

\begin{document}
\maketitle

\begin{abstract}
Recent Long-Context Language Models (LCLMs) can process hundreds of thousands of tokens in a single prompt, enabling new opportunities for knowledge-intensive multi-hop reasoning by integrating large sets of retrieved documents or, in some cases, directly all necessary information. 
However, simply feeding more documents into the context window fails to capture how evidence should be connected.
We address this gap with \emph{thought templates}, which recast reasoning as reusable thought caches, derived from prior problem solving traces, structuring how evidence is combined and guiding multi-hop inference with factual documents.
To keep these templates effective, we propose an update strategy that iteratively refines templates derived from training data through natural-language feedback.
Across diverse benchmarks and LCLM families, our approach delivers consistent gains over strong baselines in both retrieval-based and retrieval-free settings. 
Furthermore, we show that optimized templates can be distilled into smaller open-source models, demonstrating its broad applicability and transparent reasoning reuse.
% We observe the emergence of common compositional templates, suggesting that such structured reasoning can generalize across domains.
We refer to our framework as Thought Template Augmented LCLMs (\textsc{ToTAL})\footnote{Code: \url{https://github.com/starsuzi/ToTAL}.}.
\end{abstract}
\section{Introduction}
Knowledge-intensive multi-hop reasoning tasks require models to gather evidence from multiple documents, and combine it through reasoning~\cite{musique, ircot, Tang2024MultiHopRAGBR, DBLP:conf/acl/HuangVLP25}. 
These tasks are difficult because relevant evidence must not only be identified, but also be connected in a structured way, requiring knowledge-based reasoning. 
The standard solution, Retrieval-Augmented Generation (RAG), tackles this by first retrieving a small set of relevant documents and then generating an answer from them~\cite{DBLP:conf/nips/LewisPPPKGKLYR020, Adaptive-RAG}.

\begin{figure}[t]
    \centering
    \includegraphics[width=0.975\columnwidth]{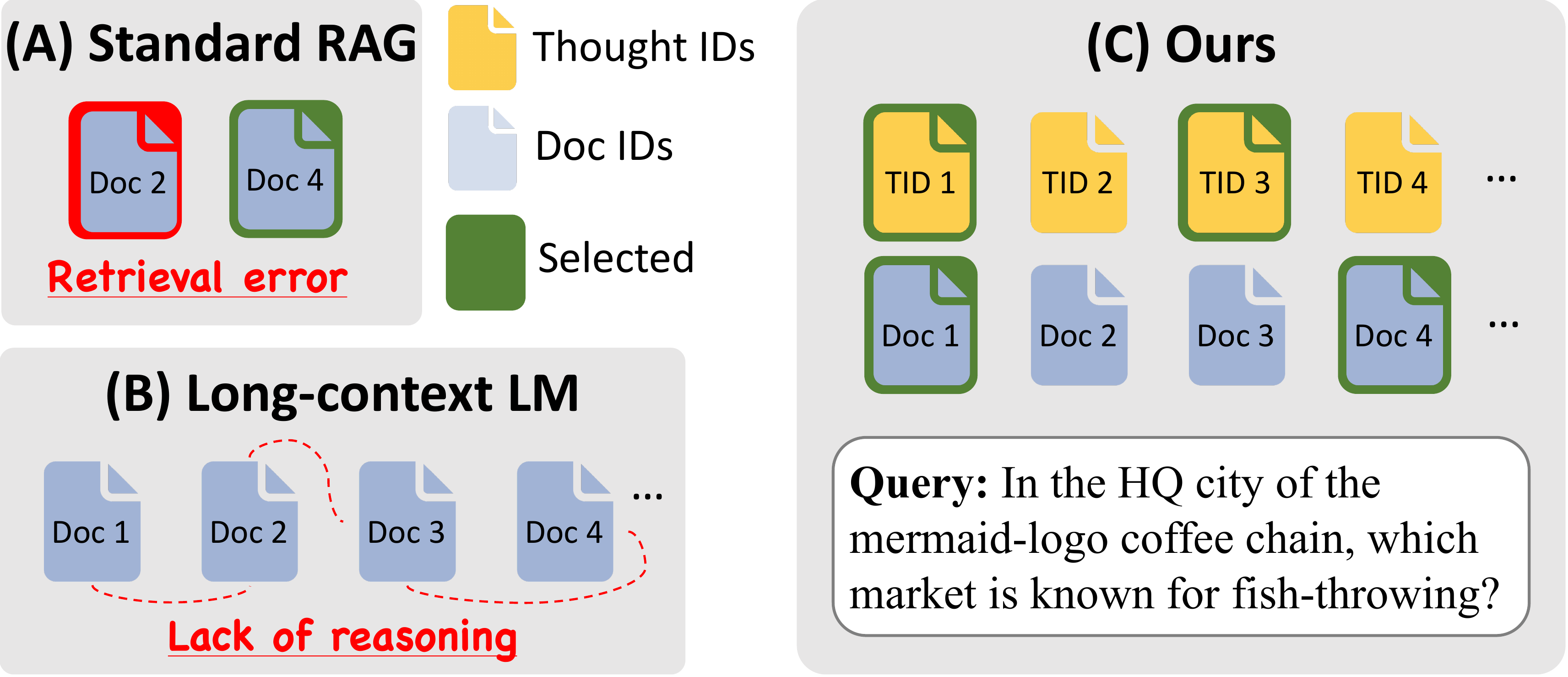}
    \vspace{-0.025in}
    \caption{\small Thoughts and facts in LCLM, compared to transitional RAG and simple stuffing in LCLM. 
    }
    \label{fig:motivation_concept}
    \vspace{-0.10in}
\end{figure}

The rise of Long-Context Language Models (LCLMs) has shifted this paradigm by enabling prompts of hundreds of thousands of tokens~\cite{Gemini, Gemini2.5, claude4, gpt5}. This advancement makes it possible to “just put everything into the prompt,” such as feeding in all retrieved documents~\cite{loft} or many in-context examples~\cite{Agarwal2024ManyShotIL, DBLP:conf/acl/BaekLGODK25}. Compared to conventional RAG, which risks cascading errors from retrieval, LCLMs support a one-step formulation that mitigates such errors, and in some domains (e.g., enterprise settings) can even absorb an entire document collection into the prompt. 
However, increasing recall with more documents alone remains insufficient, since models may struggle to connect pieces of evidence. Existing work on LCLMs has largely focused on scaling input size rather than strengthening reasoning, leaving this gap unaddressed.

While one possible direction is adopting reasoning strategies such as Chain-of-Thought~\cite{cot}, which elicit step-by-step reasoning, it remains ad-hoc and query-specific, and they are not designed to cope with the vast, document-heavy contexts enabled by LCLMs.
To address this, we introduce \textit{thought templates}: reusable reasoning patterns (or epistemic knowledge from prior experience) that act as structured scaffolds for integrating and organizing evidence in these long-context settings.
Templates act as a cache of prior reasoning behaviors capturing \textit{how to think}, while documents provide the factual content capturing \textit{what to know}.
Importantly, although the entire template set is supplied, LCLMs selectively leverage on the relevant ones for each query, thereby enabling compositional reasoning over complex evidence.

To instantiate this idea, we automatically construct templates from multi-hop QA datasets in a compositional manner, allowing LCLMs to flexibly recombine multiple templates within a single generation. Unlike prior approaches, which retrieve a single problem-specific reasoning trace~\cite{BoT, ReasonFlux}, our method enables reusability across queries. This compositional design 
also improves performance by allowing LCLMs to generalize to more complex reasoning tasks.
To further improve effectiveness, we treat thought templates as external parameters of LCLMs and refine them iteratively using natural language feedback. Feedback derived from model errors specifies how templates should be revised, functioning like a gradient update but without altering model weights.

We present a framework, Thought Template Augmented LCLMs (\textsc{ToTAL}), that equips long-context models with reusable reasoning patterns and iteratively refines them through natural language feedback.
We validate \textsc{ToTAL} on diverse knowledge-intensive datasets that require both factual grounding and multi-hop reasoning. Furthermore, we evaluate it in two settings: an idealized setup without retrieval and a more practical scenario with retrieval. 
Across both settings, thought templates consistently boost LCLM performance, and our feedback-driven update strategy yields additional gains.
These results highlight the promise of equipping LCLMs with structured reasoning patterns rather than relying solely on larger contexts.

\section{Proposed Method}
Our method is motivated by three observations: 
(1) simply increasing the number of accessible documents in LCLMs does not guarantee better reasoning; 
(2) current models often lack explicit, structured strategies for combining evidence across multiple steps; and
(3)  once distilled, such strategies can be generalized and reused across models.  
Below we introduce the necessary background and describe the design of \textsc{ToTAL}.

\subsection{Preliminaries}
We first outline the challenges and the limitations of existing paradigms of multi-hop reasoning.

\paragraph{Knowledge-intensive Multi-hop Reasoning}
Multi-hop reasoning requires gathering and integrating evidence scattered across multiple documents and composing intermediate steps for the final answer. Formally, given a query $\vq$ and a large corpus of documents $\mathcal{D} = \{\vd_1, \vd_2, \dots, \vd_N\}$, the objective is to generate the correct answer $\va$ by selecting a relevant evidence subset $\mathcal{D}_{q} \subseteq \mathcal{D}$ and chaining reasoning steps over it.

\paragraph{Retrieval-Augmented Generation (RAG)}
Conventional approaches rely on RAG: a retriever first identifies a subset of documents $\mathcal{D}_q = \texttt{Retriever}(\vq, \mathcal{D})$, and then a Language Model (LM) generates an answer conditioned on both $\vq$ and $\mathcal{D}_q$, denoted as $\va = \texttt{LM}(\vq, \mathcal{D}_q)$. 
Since earlier LMs were limited by context length, retrieval quality was crucial: poor retrieval caused cascading errors by omitting essential evidence. 

\begin{figure*}
    \centering
    \includegraphics[width=0.99\linewidth]{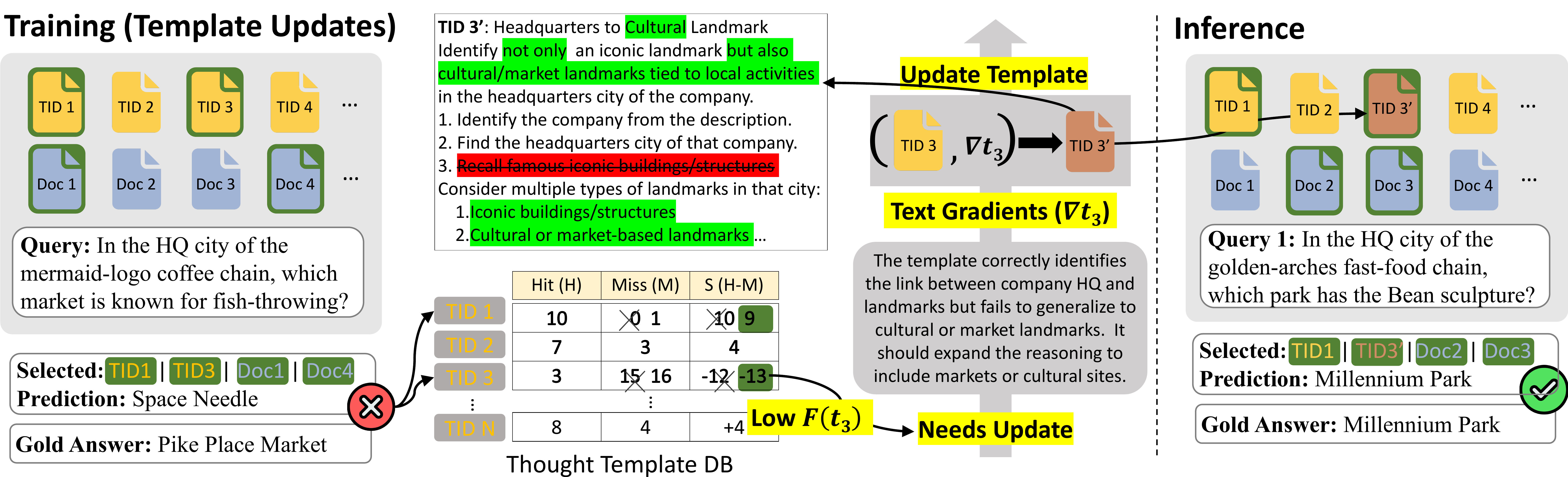}
    \vspace{-0.1in}
    \caption{\small 
    Illustration of training and inference stages for template updates. Low-performing templates are identified via hit/miss statistics and refined with textual gradient feedback, enabling improved performance on new queries during inference.
    }
    \label{fig:fig_concept_method}
    \vspace{-0.1in}
\end{figure*}

\paragraph{Long-Context Language Models (LCLMs)}
Recent LCLMs can process even millions of tokens in a single prompt, thereby allowing the direct inclusion of large evidence sets (or entire corpora) into the context: $\va = \texttt{LCLM}(\vq, \mathcal{D})$.
Alternatively, retrieval still can be used to select a much larger set of documents than before, $\mathcal{D}_q^{*} = \texttt{Retriever}(\vq, \mathcal{D})$, where $|\mathcal{D}_q| \ll |\mathcal{D}_q^{*}|$.
Thus, LCLM supports two regimes: inserting full corpus $\mathcal{D}$, or large retrieved subset $\mathcal{D}_{\texttt{large}}$, where $\mathcal{D}_{\texttt{large}} \in \{\mathcal{D}, \mathcal{D}_q^{*}\}$.
However, simply scaling document access is insufficient: the bottleneck now lies in how to \textit{structure} and \textit{reuse} reasoning over abundant knowledge.
At the same time, finetuning LCLM to explicitly learn long reasoning chains is often infeasible due to their high cost and limited accessibility.

\subsection{Thought Template Augmented LCLMs}
To bridge these gaps, we introduce \textsc{ToTAL}, a framework that enables better reasoning in large document contexts \textit{without any model finetuning} by leveraging \textit{thought templates} -- structured thought processes built from training data and refined iteratively through textual gradient feedback (Figure~\ref{fig:fig_concept_method}).
These updated templates then guide the LCLM in organizing evidence and performing multi-step reasoning more effectively during inference.

\paragraph{Thought Templates}
A \textit{thought template} is a reusable high-level reasoning pattern, distilled from prior problem-solving.
Each template provides a structured outline of intermediate steps that can be instantiated for new queries.
Formally, let $\mathcal{T} = \{\vt_1, \vt_2, \dots, \vt_m\}$ denote the set of templates.
At inference, the LCLM is conditioned on both the query $\vq$, large evidence set $\mathcal{D}_{\texttt{large}}$, and templates: 
\[
\hat{\va} = \texttt{LCLM}(\vq, \mathcal{T}, \mathcal{D}_{\texttt{large}})
\]

\paragraph{Template Construction}
To build the initial template set $\mathcal{T}$, we prompt an LCLM to generate templates, conditioned on the training queries $\vq_{\texttt{train}}$, their gold answers $\va_{\texttt{train}}$, and optionally solution paths $\vs_{\texttt{train}}$  from the training set: 
\[\vt_i = \texttt{LCLM}(\vq_{\texttt{train}}, \va_{\texttt{train}}, [\vs_{\texttt{train}}])\]
This procedure is inspired by \citet{ReasonFlux}, who also derive templates using LLMs. However, instead of capturing full example-specific solutions, we decompose them into sub-templates that are reusable across queries and generalize more effectively.
At inference, the model selectively applies and composes relevant templates from $\mathcal{T}$ with the query and supporting documents.
Below shows an example of a thought template $\vt_3$ generated from the following template construction process.

\begin{tcolorbox}
[colback=gray!5, colframe=gray!50, boxrule=0.5pt, arc=2mm, left=4pt, right=4pt, top=3pt, bottom=3pt]
\noindent\textbf{TID 3:} Headquarters to Landmark\\[2pt]
Identify an iconic landmark in the headquarters city of the company.\\
1. Identify the company from the description.\\
2. Find the headquarters city of that company.
3. Recall famous iconic buildings/structures ...
\end{tcolorbox}

\paragraph{Template Update Strategy}
Initial templates may be noisy or suboptimal.
Thus, we iteratively refine templates $\vt_i$ using natural-language feedback as a \textit{surrogate gradient}.
We first assign each template $\vt_i$ an explicit performance score $F(\vt_i)$ to compute which reasoning patterns contribute positively or negatively to model outputs.
Specifically, for each $\vq_{\texttt{train}}$ with  $\va_{\texttt{train}}$, we obtain the model's prediction $\hat{\va}_{\texttt{train}} = \texttt{LCLM}(\vq_{\texttt{train}}, \mathcal{T}, \mathcal{D}_{\texttt{large}})$.
Then, $\vt_i$ is assigned an aggregated score as:
\[
F(\vt_i) = \sum_{\vq_{\texttt{train}}} f_i(\vq_{\texttt{train}}),
\]
where $f_i(\vq_{\texttt{train}})$ measures the performance of $\vt_i$ on $\vq_{\texttt{train}}$ by comparing $\hat{\va}_{\texttt{train}}$ with $\va_{\texttt{train}}$ (e.g., using metrics such as exact match or F1 in QA tasks). Importantly, $f_i(\vq_{\texttt{train}})$ is computed only for queries where $\vt_i$ is actually selected.
Templates with scores below a threshold $F(\vt_i) < \tau$\footnote{$\tau$ denoting a threshold selected with the validation set} are identified as low-performing and selected for refinement (e.g., TID 3 in the template database in Figure \ref{fig:fig_concept_method}).
This enables targeted refinement, updating only low-performing templates, while maintaining the stability of well-performing ones.

For each low-performing template, another LM analyzes its failure cases by comparing the query $\vq_{\texttt{train}}$, its prediction $\hat{\va}_{\texttt{train}}$, the gold answer $\va_{\texttt{train}}$, and the applied template $\vt_i$, and produces a natural-language ``textual gradient'' feedback:
\[
\nabla \vt_i = \texttt{LM}_{\texttt{Feedback}}(\vq_{\texttt{train}}, \hat{\va}_{\texttt{train}}, \va_{\texttt{train}}, \vt_i)
\]
Below is an example feedback $\nabla \vt_3$.
\begin{tcolorbox}[colback=gray!5, colframe=gray!50, 
                  boxrule=0.5pt, arc=2mm, left=4pt, right=4pt, top=3pt, bottom=3pt]
\noindent \textbf{$\nabla$ TID 3:} The template correctly identifies the link between company HQ and landmarks but fails to generalize cultural or market landmarks. It should expand the reasoning to include..
\end{tcolorbox}

\noindent This textual gradient is accompanied by a discrete decision indicating the appropriate update action:
\[
d_i \in \{\textsc{Keep}, \textsc{Fix}, \textsc{Add}, \textsc{Discard}\}
\]
For \textsc{Keep}, the template remains unchanged, while for \textsc{Discard}, the template is removed. For \textsc{Fix} and \textsc{Add}, $\nabla \vt_i$ is passed to another LM:
\[
\vt_i' = \texttt{LM}_{\texttt{update}}(\vt_i, \nabla \vt_i)
\]
This iterative refinement process progressively improves the template set $\mathcal{T}$, which is subsequently used during inference to guide reasoning.
The updated $\vt_3'$ updated from $\vt_3$ looks like below:
\begin{tcolorbox}[colback=gray!5, colframe=gray!50, 
                  boxrule=0.5pt, arc=2mm, left=4pt, right=4pt, top=3pt, bottom=3pt]
\noindent\textbf{TID 3$'$:} Headquarters to \textcolor{green!60!black}{Cultural} Landmark\\[2pt]
Identify \textcolor{green!60!black}{not only} an iconic landmark \textcolor{green!60!black}{but also cultural/market landmarks tied to local activities} in the headquarters city of the company.\\
1. Identify the company from the description.\\
2. Find the headquarters city of that company.
3. \textcolor{red}{\strut \sout{Recall famous iconic buildings/structures}}  ...
\end{tcolorbox}

\section{Experimental Setup}

\begin{table*}[t!]
\caption{\small Main results on four multi-hop reasoning datasets under the LCLM setting.
We report F1 on MuSiQue, CRAG, and FanOutQA, Accuracy on Housing QA, and the overall Average.
The best results are highlighted in \textbf{bold}.}
\vspace{-0.1in}
\label{tab:main}
\small
\centering
\renewcommand{\arraystretch}{1.0}
\resizebox{\linewidth}{!}{%
\begin{tabular}{l l c c c c >{\columncolor{gray!10}}c}
\toprule
& \textbf{Methods} & \textbf{MuSiQue (F1)} & \textbf{CRAG (F1)} & \textbf{FanOutQA (F1)} & \textbf{Housing QA (Acc.)} & \textbf{Average} \\
\midrule
\midrule
% ================= Claude =================
\multirow{5}{*}{\rotatebox{90}{\textbf{Claude}}}
& \textbf{\textsc{Naïve}}      & 27.57 {\textsmaller{± 0.27}} & 20.49 {\textsmaller{± 1.02}} & 46.72 {\textsmaller{± 1.15}} & 60.33 {\textsmaller{± 2.08}} & 38.78 \\
& \textbf{\textsc{CoT}}        & 28.10 {\textsmaller{± 0.91}} & 20.32 {\textsmaller{± 1.29}} & 45.54 {\textsmaller{± 0.13}} & 57.67 {\textsmaller{± 2.08}} & 37.90 \\
& \textbf{\textsc{CiC}}        & 63.87 {\textsmaller{± 0.91}} & 17.32 {\textsmaller{± 0.57}} & 63.74 {\textsmaller{± 1.50}} & 71.67 {\textsmaller{± 0.58}} & 54.15 \\
& \textbf{\textsc{CiC + CoT}}  & 65.07 {\textsmaller{± 0.16}} & 18.86 {\textsmaller{± 0.01}} & 66.29 {\textsmaller{± 0.53}} & 75.00 {\textsmaller{± 2.00}} & 56.30 \\
& \textbf{\textsc{ToTAL} (Ours)} & \textbf{73.30 {\textsmaller{± 1.24}}} & \textbf{30.08 {\textsmaller{± 0.83}}} & \textbf{69.99 {\textsmaller{± 1.61}}} & \textbf{82.67 {\textsmaller{± 0.58}}} & \textbf{64.01} \\

\midrule
\midrule

% ================= Gemini =================
\multirow{5}{*}{\rotatebox{90}{\textbf{Gemini}}}
& \textbf{\textsc{Naïve}}      & 25.48 {\textsmaller{± 1.85}} & 22.03 {\textsmaller{± 1.31}} & 46.54 {\textsmaller{± 1.85}} & 58.00 {\textsmaller{± 2.00}} & 38.01 \\
& \textbf{\textsc{CoT}}        & 23.03 {\textsmaller{± 0.80}} & 24.62 {\textsmaller{± 0.74}} & 43.54 {\textsmaller{± 1.54}} & 58.67 {\textsmaller{± 2.89}} & 37.46 \\
& \textbf{\textsc{CiC}}        & 66.54 {\textsmaller{± 1.10}} & 25.45 {\textsmaller{± 0.68}} & 66.44 {\textsmaller{± 1.33}} & 68.33 {\textsmaller{± 1.15}} & 56.69 \\
& \textbf{\textsc{CiC + CoT}}  & 67.17 {\textsmaller{± 1.01}} & 25.77 {\textsmaller{± 0.30}} & 66.97 {\textsmaller{± 0.48}} & 70.33 {\textsmaller{± 0.58}} & 57.56 \\
& \textbf{\textsc{ToTAL} (Ours)} & \textbf{72.86 {\textsmaller{± 0.71}}} & \textbf{27.71 {\textsmaller{± 0.51}}} & \textbf{71.84 {\textsmaller{± 0.62}}} & \textbf{74.33 {\textsmaller{± 0.58}}} & \textbf{61.68} \\

\midrule
\midrule

% ================= GPT =================
\multirow{5}{*}{\rotatebox{90}{\textbf{GPT}}}
& \textbf{\textsc{Naïve}}      & 32.43 {\textsmaller{± 0.67}} & 25.73 {\textsmaller{± 0.62}} & 48.77 {\textsmaller{± 1.41}} & 60.33 {\textsmaller{± 0.58}} & 41.81 \\
& \textbf{\textsc{CoT}}        & 32.39 {\textsmaller{± 0.15}} & 23.24 {\textsmaller{± 0.89}} & 49.09 {\textsmaller{± 0.97}} & 61.33 {\textsmaller{± 0.58}} & 41.51 \\
& \textbf{\textsc{CiC}}        & 63.79 {\textsmaller{± 0.95}} & 22.12 {\textsmaller{± 0.69}} & 63.39 {\textsmaller{± 1.23}} & 64.33 {\textsmaller{± 0.58}} & 52.50 \\
& \textbf{\textsc{CiC + CoT}}  & 65.11 {\textsmaller{± 0.44}} & 21.72 {\textsmaller{± 0.75}} & 66.35 {\textsmaller{± 0.62}} & 66.00 {\textsmaller{± 1.00}} & 54.79 \\
& \textbf{\textsc{ToTAL} (Ours)} & \textbf{66.38 {\textsmaller{± 0.13}}} & \textbf{26.31 {\textsmaller{± 0.74}}} & \textbf{69.07 {\textsmaller{± 1.83}}} & \textbf{70.00 {\textsmaller{± 1.00}}} & \textbf{57.94} \\
\bottomrule
\end{tabular}
}
\vspace{-0.1in}
\end{table*}

\begin{table}[t]
\centering\small
\caption{RAG results of LCLMs with retrieved documents.}
\vspace{-0.1in}
\label{tab:rag_results}
\renewcommand{\arraystretch}{1.0}
\renewcommand{\tabcolsep}{2mm}
\resizebox{\linewidth}{!}{%
\begin{tabular}{lcccc}
    \toprule
    \textbf{Methods} & \textbf{MSQ} & \textbf{CRAG} & \textbf{FOQA} & \textbf{HQA} \\
    \midrule
    \midrule
    \textbf{\textsc{CiC}} & 41.63 & 13.10 & 26.57 & 70.00  \\
    \textbf{\textsc{ToTAL} (Ours)} & \textbf{47.90} & \textbf{19.87} & \textbf{32.16} & \textbf{76.50} \\
    \bottomrule
\end{tabular}
}
\vspace{-0.05in}
\end{table}

\subsection{Datasets}
We evaluate \textsc{ToTAL} on four challenging multi-hop QA benchmarks: MuSiQue~\cite{musique}, CRAG~\cite{crag}, FanOutQA~\cite{FanOutQA}, and Housing QA~\cite{housingqa}. MuSiQue requires reasoning over multiple passages and is widely used for evaluating multi-hop question answering. 
CRAG focuses on diverse and dynamic queries, going beyond traditional datasets by incorporating less popular topics and more complex reasoning types.
FanOutQA consists of long-context Wikipedia documents.
Housing QA evaluates domain-specific legal queries that require retrieving and reasoning over statutory texts.

\subsection{Baseline Models}
We compare \textsc{ToTAL} against four representative baselines, all of which use LCLMs as base models: 

\begin{itemize}[leftmargin=1.5em, itemsep=0.15pt, topsep=0.15pt]
    \item \textbf{\textsc{Naïve:}} Directly generates answers from the query without any auxiliary context.  
    \item \textbf{\textsc{Chain-of-Thought (CoT)}}~\cite{zeroshotcot}: A prompting-based reasoning approach using the phrase ``Let's think step by step.''
    \item \textbf{\textsc{Corpus-in-Context (CiC)}}~\cite{loft}: Leverages the extended context window of LCLMs by inserting the entire documents directly into the prompt.  
    \item \textbf{\textsc{CiC + CoT:}} Combines \textsc{CiC} with \textsc{CoT}, aiming to jointly utilize large-context access and explicit reasoning cues.  
\end{itemize}

\textsc{ToTAL} differs from these baselines by introducing structured, reusable reasoning patterns (\emph{thought templates}) that guide LCLMs to organize and apply evidence effectively, without additional fine-tuning.

\subsection{Implementation Details}
We evaluate across a diverse suite of LCLMs, including proprietary frontier models, Claude-Sonnet 4~\cite{claude4}, and Gemini 2.5 Flash~\cite{Gemini2.5}, and GPT-4.1~\cite{gpt41} as well as open-source LLMs such as OSS (120B)~\cite{oss} and DeepSeek-R1~\cite{deepseek-r1}.  
For retrieval-based settings, we employ BM25~\cite{bm25} as the retriever.
We adopt standard QA metrics tailored to each dataset: F1 score for MuSiQue, CRAG, and FanOutQA, and Accuracy for Housing QA with binary outputs, and use the same metrics to compute template scores.
Unless stated otherwise, we primarily use Claude on MuSiQue for analyses.
We provide the prompts for template construction and update in Figures~\ref{fig:prompt_template_construction_compositional}, \ref{fig:prompt_textgrad_feedback}, and \ref{fig:prompt_textgrad_edit}.

\subsection{Data Processing}
For MuSiQue, we use the 128k-token version from the LOFT benchmark~\cite{loft}, and apply a similar preprocessing procedure to the other datasets, including matching the number of test queries.
For CRAG, we focus on the \emph{Multi-hop} and \emph{Post-processing heavy} categories to target complex reasoning cases.  
For both CRAG and Housing QA, we construct corpora by aggregating all relevant snippets or statutes for each query, capped at 128k tokens.  
For FanOutQA, since its context units are full Wikipedia pages, we build a query-specific corpus containing only documents relevant to each question, also truncated to 128k tokens. 

For the retrieval setting, we first construct 1M-token corpora for MuSiQue, CRAG, and Housing QA, and then subsample them to match the 128k-token budget 
(800 documents out of 6,650 for MuSiQue, 300 out of 2,307 for CRAG, and 480 out of 5,924 for Housing QA). 
For FanOutQA, we use all 2,142 documents from the original corpus as the retrieval corpus, and then retrieve 5 documents.

Regarding the template design, we construct the initial template set by sampling 50 QA pairs from the training data, ensuring no overlap with the test queries. For the template update strategy, we also use another subset of the training samples and determine the threshold~$\tau$ on the validation set.

\section{Results and Analyses}

\subsection{Main Results}
Table~\ref{tab:main} presents the performance across all benchmarks. 
Although recent LCLMs demonstrate strong capabilities, they still struggle with complex and knowledge-intensive multi-hop queries.
This is reflected in the performance of the \textsc{Naïve} baseline, which lacks access to external facts and relies on the internalized knowledge.
Similarly, \textsc{Chain-of-Thought (CoT)} prompting yields only marginal improvements, suggesting that explicitly eliciting step-by-step reasoning alone is insufficient for multi-hop knowledge integration.
\textsc{Corpus-in-Context (CiC)} improves performance by leveraging the extended context window of LCLMs to include the entire corpus in the prompt.  
However, its gains remain limited since it treats the task as evidence aggregation rather than reasoning composition. 
Even when combined with CoT (\textsc{CiC + CoT}), improvements are modest.
In contrast, \textsc{ToTAL} introduces implicit reasoning structure through reusable templates, \textit{consistently outperforming all baselines across datasets}.  
This highlights the value of guiding LCLMs with structured reasoning patterns rather than relying solely on surface-level prompting strategies.

\begin{figure}
    \centering
    \vspace{-0.05in}
    \includegraphics[width=0.975\linewidth]{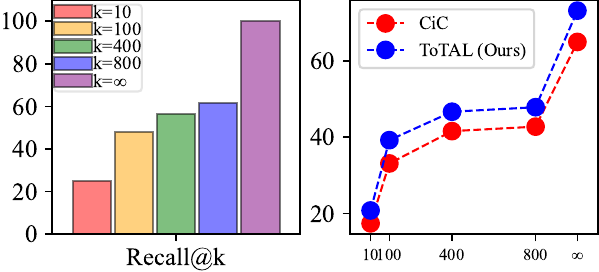}
    \vspace{-0.10in}
    \caption{\small RAG results on MuSiQue, showing retrieval recall at different $k$ values (left) and QA performance (F1) (right).
    }
    \label{fig:fig_rag_recall}
    \vspace{-0.10in}
\end{figure}

\subsection{When Partial Context Given}
While LCLMs are capable of handling large contexts, retrieval becomes essential when the full corpus cannot be included. 
We evaluate this retrieval-augmented scenario by comparing \textsc{ToTAL} with \textsc{CiC}, with both models given the same retrieved documents under our full-context budget.
Table~\ref{tab:rag_results} shows that \textsc{ToTAL} consistently outperforms \textsc{CiC}, demonstrating that reasoning templates provide complementary advantages even in retrieval settings. 
To further examine this scenario, we vary the number of retrieved documents ($k$) to emulate more realistic retrieval-augmented scenarios in Figure~\ref{fig:fig_rag_recall}.
As the number of retrieved documents increases, both retrieval recall and QA performance improve, confirming that \textit{long-context models benefit from expanded evidence access}.
However, when compared with the idealized setting where all documents are available to the LCLM ($k = \infty$), retrieval still imposes a bottleneck.  
Importantly, our template-based approach consistently enhances performance across all retrieval sizes, demonstrating its robustness and adaptability.  
As LCLMs continue to scale in context length, such structured reasoning strategies are expected to further amplify their benefits.

\begin{figure}
    \centering
    \includegraphics[width=0.975\linewidth]{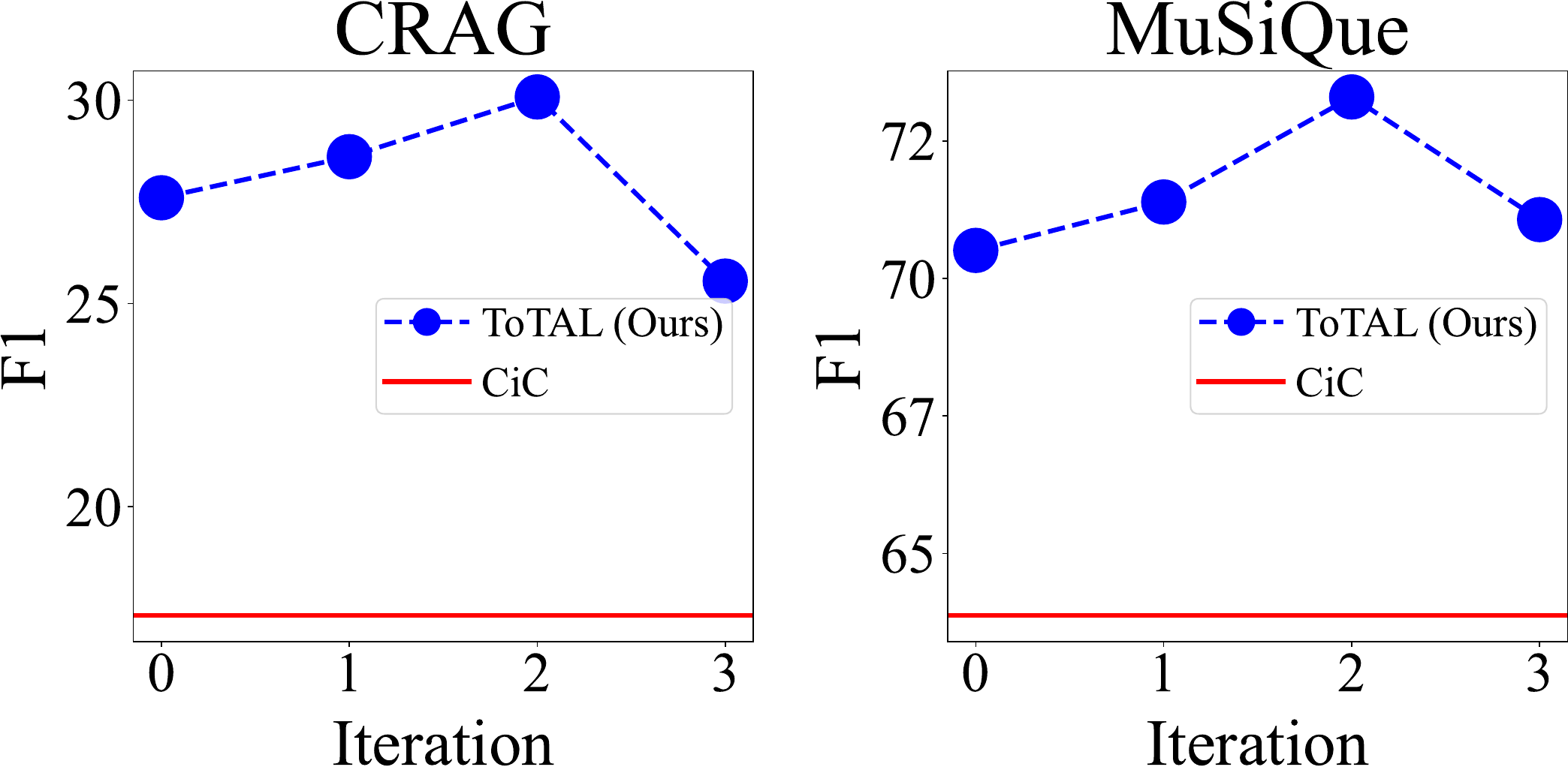}
    \vspace{-0.075in}
    \caption{\small Iteration results of updates on CRAG and MuSiQue.}
    \label{fig:fig_iteration}
    \vspace{0in}
\end{figure}
\begin{table}[t!]
\caption{\small Transferability of templates across LCLMs, where templates generated by GPT and Gemini are applied to Claude.}
\vspace{-0.075in}
\label{tab:transfer_frontier}
\small
\centering
\resizebox{0.475\textwidth}{!}{
\renewcommand{\arraystretch}{1.175}
\renewcommand{\tabcolsep}{2.8mm}
\begin{tabular}{l c c}
\toprule
\textbf{Methods} & \textbf{Source $\rightarrow$ Target} & \textbf{F1} \\
\midrule
\midrule
\multirowcell{1}[-0.0ex][l]{\textbf{\textsc{CiC}}} 
& -- & 63.87 \\
\noalign{\vskip 0.25ex}\cdashline{1-3}\noalign{\vskip 0.75ex}
\multirowcell{1}[-0.0ex][l]{\textbf{\textsc{ToTAL} (Ours)}} 
& Gemini $\rightarrow$ Claude & \textbf{70.94} \\
\multirowcell{1}[-0.0ex][l]{} 
& GPT $\rightarrow$ Claude & \textbf{70.11} \\
\bottomrule
\end{tabular}
}
\vspace{-0.025in}
\end{table}

\subsection{Effectiveness of Template Update Strategy}
To assess the contribution of the template update strategy, we report ablation results in Figure~\ref{fig:fig_iteration}.  
The results show clear performance gains when applying iterative updates over the initial template set.  
\textit{Refining low-performing templates via feedback substantially enhances reasoning accuracy}, validating our design choice of using natural-language feedback as a surrogate optimization signal.  
Even without updates, the initial template set already outperforms \textsc{CiC}, indicating that structured reasoning guidance itself contributes significant benefits.
Performance plateauing around the second iteration reflects a diminishing returns effect, commonly observed in conventional ML, suggesting that the updates have effectively converged. We further show the updates across iterations in Appendix~\ref{appen:template_iteration_operation}.

\begin{figure}[t]
    \centering
    \includegraphics[width=0.975\linewidth]{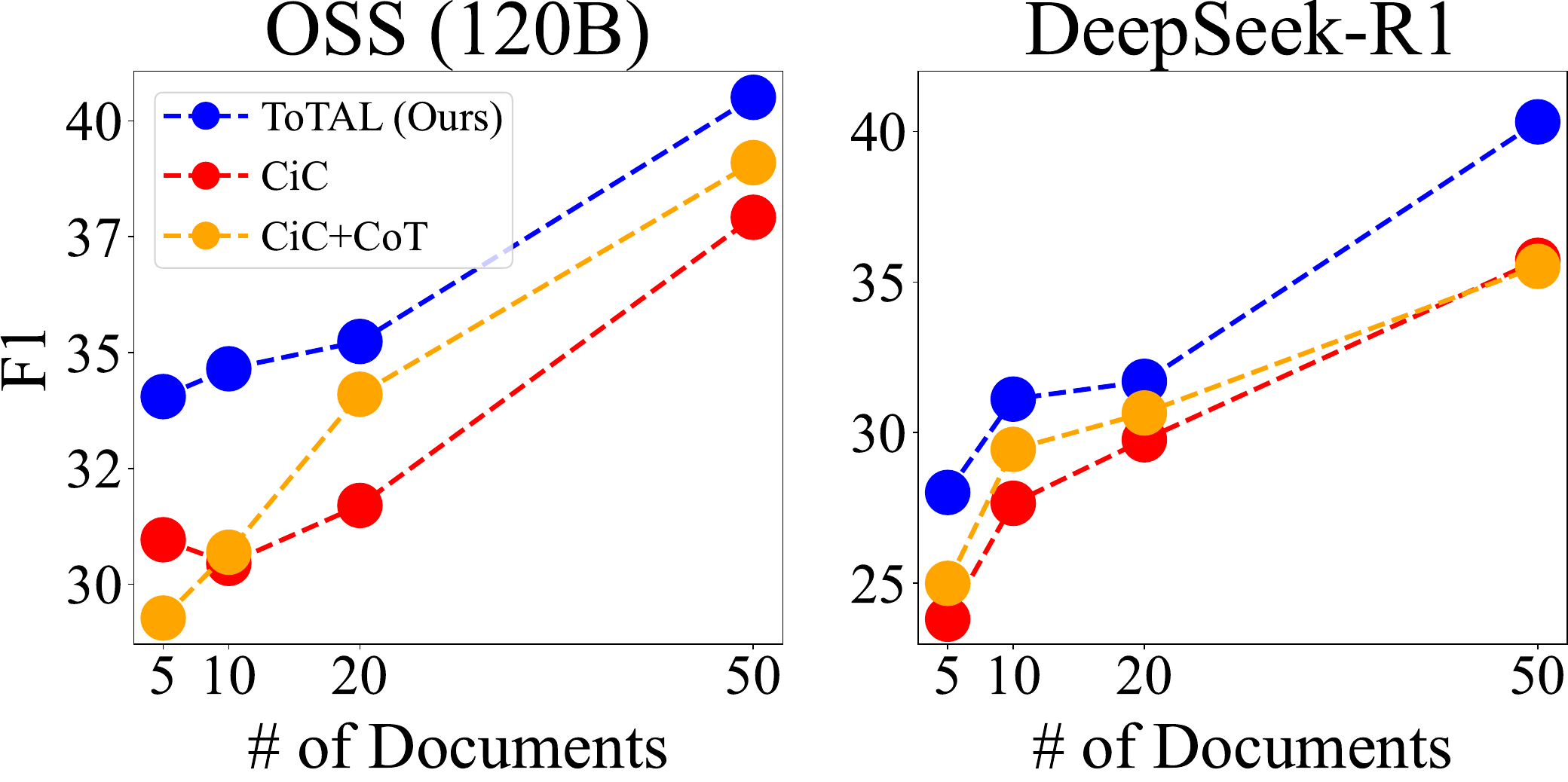}
    \vspace{-0.075in}
    \caption{\small Generalization of templates to open-source models.}
    \label{fig:fig_open_models}
    \vspace{0in}
\end{figure}

\subsection{Transferability of Templates} 
We evaluate generalization by testing template transfer across frontier models and applying templates from Claude to open-source LLMs.
Table~\ref{tab:transfer_frontier} shows that templates from one frontier LCLM can be effectively applied to others.
Similarly, Figure~\ref{fig:fig_open_models} shows that these templates also transfer well to open-source models, yielding consistent improvements over baselines even under shorter context windows and retrieval-augmented settings.
The gains remain stable across varying top-$k$ values, underscoring that \textit{thought templates encode model-agnostic reasoning structures} that generalize across architectures and retrieval conditions.

\begin{table}[t!]
\caption{\small Performance when using templates generated by open-source models versus distilled from a frontier model.}
\vspace{-0.075in}
\label{tab:small_templates}
\small
\centering
\resizebox{0.475\textwidth}{!}{
\renewcommand{\arraystretch}{1.0}
\renewcommand{\tabcolsep}{3mm}
\begin{tabular}{lcc}
\toprule
\textbf{Methods} & \textbf{OSS} & \textbf{DeepSeek-R1} \\
\midrule
\midrule

\multirowcell{1}[-0.0ex][l]{\textbf{\textsc{Naïve}}}
& 17.27 & 14.98 \\

\noalign{\vskip 0.25ex}\cdashline{1-3}\noalign{\vskip 0.75ex}

\multirowcell{1}[-0.0ex][l]{\textbf{\textsc{CiC}}} 
& 30.45 & 27.65 \\

\multirowcell{1}[-0.0ex][l]{\textbf{\textsc{ToTAL} w/ Open}} 
& 32.44 & 29.53 \\

\multirowcell{1}[-0.0ex][l]{\textbf{\textsc{ToTAL} w/ Distilled}} 
& \textbf{34.65} & \textbf{31.11} \\

\bottomrule
\end{tabular}
}
\vspace{-0.0in}
\end{table}

\paragraph{Templates Generated by Open-Source LLMs.}
We further investigate whether templates can be generated and refined entirely by open-source models.  
In Table~\ref{tab:small_templates}, templates produced and updated by the same open-source model already surpass the \textsc{CiC} baseline, confirming the feasibility of a fully open pipeline.  
While templates derived from frontier LCLMs achieve higher performance, results suggest that template quality scales with model capacity, yet open-source systems can still produce competitive and practical reasoning patterns.

\begin{table}[t!]
\centering\small
\caption{Cross-dataset results from the close clusters on CRAG with templates from FanOutQA or Housing QA.}
\vspace{-0.075in}
\label{tab:cross_templates}
\renewcommand{\arraystretch}{0.9}
\renewcommand{\tabcolsep}{4mm}
\resizebox{\linewidth}{!}{%
\begin{tabular}{lc}
    \toprule
    \textbf{Methods} & \textbf{F1} \\
    \midrule
    \midrule
    \textbf{\textsc{CiC}} & 17.32 \\
    \noalign{\vskip 0.25ex}\cdashline{1-2}\noalign{\vskip 0.75ex}
    \textbf{\textsc{ToTAL} (with templates from Housing QA)} & 18.29  \\
    \textbf{\textsc{ToTAL} (with templates from FanOutQA} & 21.74 \\
    \textbf{\textsc{ToTAL} (Ours)} & \textbf{30.08} \\
    \bottomrule
\end{tabular}
}
%\vspace{-0.025in}
\end{table}

\paragraph{Cross-Dataset Transferability of Templates.}
While our reusability claim is made at the task level rather than across domains, we further evaluate cross-dataset transfer to examine template reusability, beyond model-level transferability. As shown in Table~\ref{tab:cross_templates}, templates transferred from a closely related dataset (FanOutQA) still provide substantial improvements over \textsc{CiC}. In contrast, templates transferred from a more distant domain (Housing QA) yield weaker performance, indicating that templates capture domain-specific reasoning structures. We believe that investigating reusability beyond the task level across more diverse domains is an interesting direction, which we leave for future work.

\begin{table}[t!]
\centering\small
\caption{Results without compositionality and with oracle.}
\vspace{-0.075in}
\label{tab:oracle_templates}
\renewcommand{\arraystretch}{0.9}
\renewcommand{\tabcolsep}{4mm}
\resizebox{\linewidth}{!}{%
\begin{tabular}{lc}
    \toprule
    \textbf{Methods} & \textbf{F1} \\
    \midrule
    \midrule
    \textbf{\textsc{CiC}} & 63.87 \\
    \noalign{\vskip 0.25ex}\cdashline{1-2}\noalign{\vskip 0.75ex}
    \textbf{\textsc{ToTAL} (Ours)} & 73.30  \\
    \textbf{\textsc{ToTAL} w/o Compositional Templates} & 67.80 \\
    \textbf{\textsc{ToTAL} w/ Oracle Templates} & \textbf{78.49} \\
    \bottomrule
\end{tabular}
}
\vspace{0.01in}
\end{table}

\subsection{Impact of Template Quality}
To get a more generalizable template, we design it compositionally, decomposing reasoning into multiple sub-templates rather than a single holistic one encompassing all steps.  
Table~\ref{tab:oracle_templates} shows that removing compositionality causes a measurable performance drop, confirming that smaller, modular templates promote better generalization across queries.  
In addition, to estimate the upper bound of our framework, we evaluate \textbf{oracle} templates constructed directly from \textit{test queries}.  
The results show that oracle templates achieve substantially higher scores, representing the potential performance ceiling achievable with perfect template design.  
The performance gap between oracle and learned templates highlights interesting and promising directions for future work, such as automatic template search and meta-learning strategies for reasoning refinement.

\begin{figure}[t]
    \centering
    \includegraphics[width=0.975\linewidth]{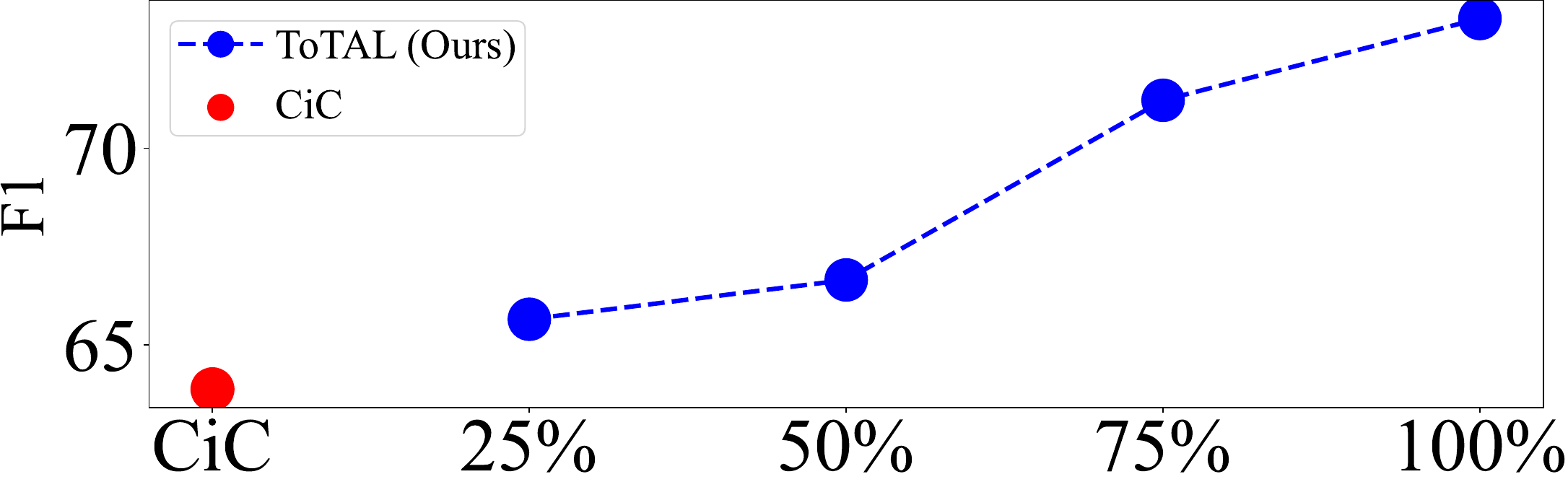}
    \vspace{-0.075in}
    \caption{\small Varying the percentage of templates on MuSiQue.}
    \label{fig:fig_num_templates}
    \vspace{-0.025in}
\end{figure}

\subsection{Impact of Template Quantity}
We also study how the number of templates affects performance by sampling different proportions (bottom 25\%, 50\%, and 75\%) of the template pool based on their scores.  
As shown in Figure~\ref{fig:fig_num_templates}, performance remains competitive even with only 25\% of the templates, and continues to improve as more templates are included.  
This suggests that high-scoring templates encode broadly reusable reasoning patterns, enabling strong performance even when the template set is reduced, although including the full set yields the best overall results.

\begin{figure}[t]
    \centering
    \includegraphics[width=0.975\linewidth]{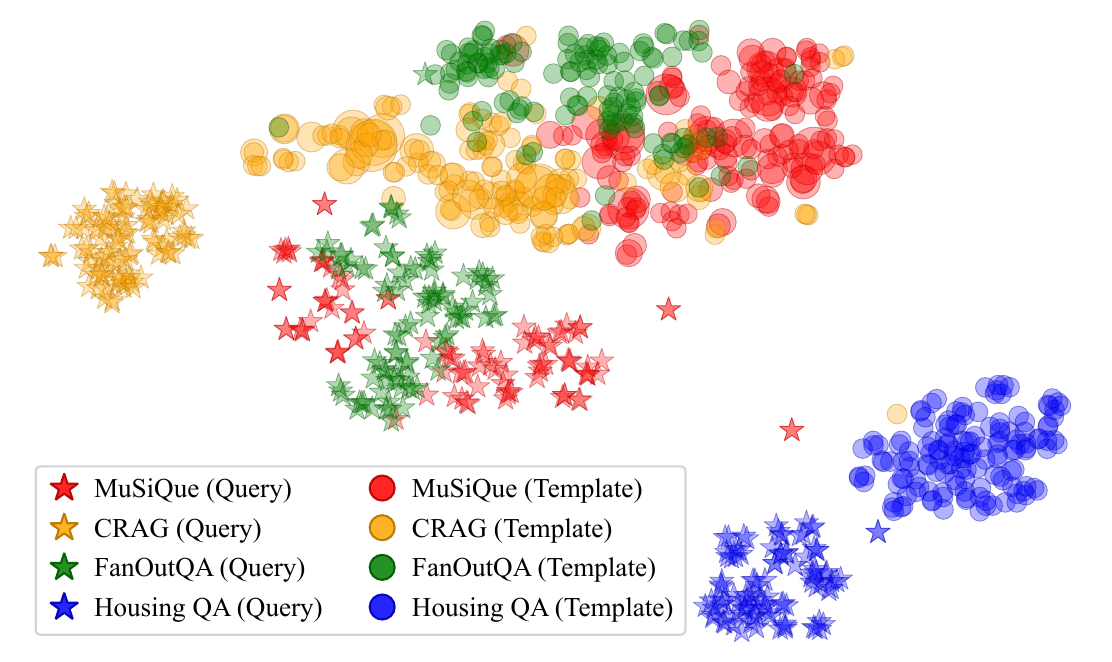}
   \vspace{-0.075in}
    \caption{\small TSNE~\cite{JMLR:v9:vandermaaten08a} visualization of the queries and templates, using embeddings from Sentence-BERT~\cite{DBLP:conf/emnlp/ReimersG19}.}
    \label{fig:tsne_queries_vs_templates_all}
    \vspace{-0.1in}
\end{figure}

\subsection{Template Analyses Beyond Performance}

\paragraph{Template–Query Clustering.}
As illustrated in Figure~\ref{fig:tsne_queries_vs_templates_all}, queries and their associated templates form coherent clusters, indicating that templates capture dataset-specific reasoning patterns aligned with the semantic structure of queries.  
Notably, the legal-domain dataset (Housing QA) appears as a clearly distinct cluster, with its templates also separated from others.  
This separation suggests that templates not only reflect domain-specific reasoning structures, but also facilitate tight coupling between queries and reusable reasoning routines.

\begin{figure*}[t]
    \centering
    \begin{minipage}{0.25\linewidth}
        \centering
        \includegraphics[width=\linewidth]{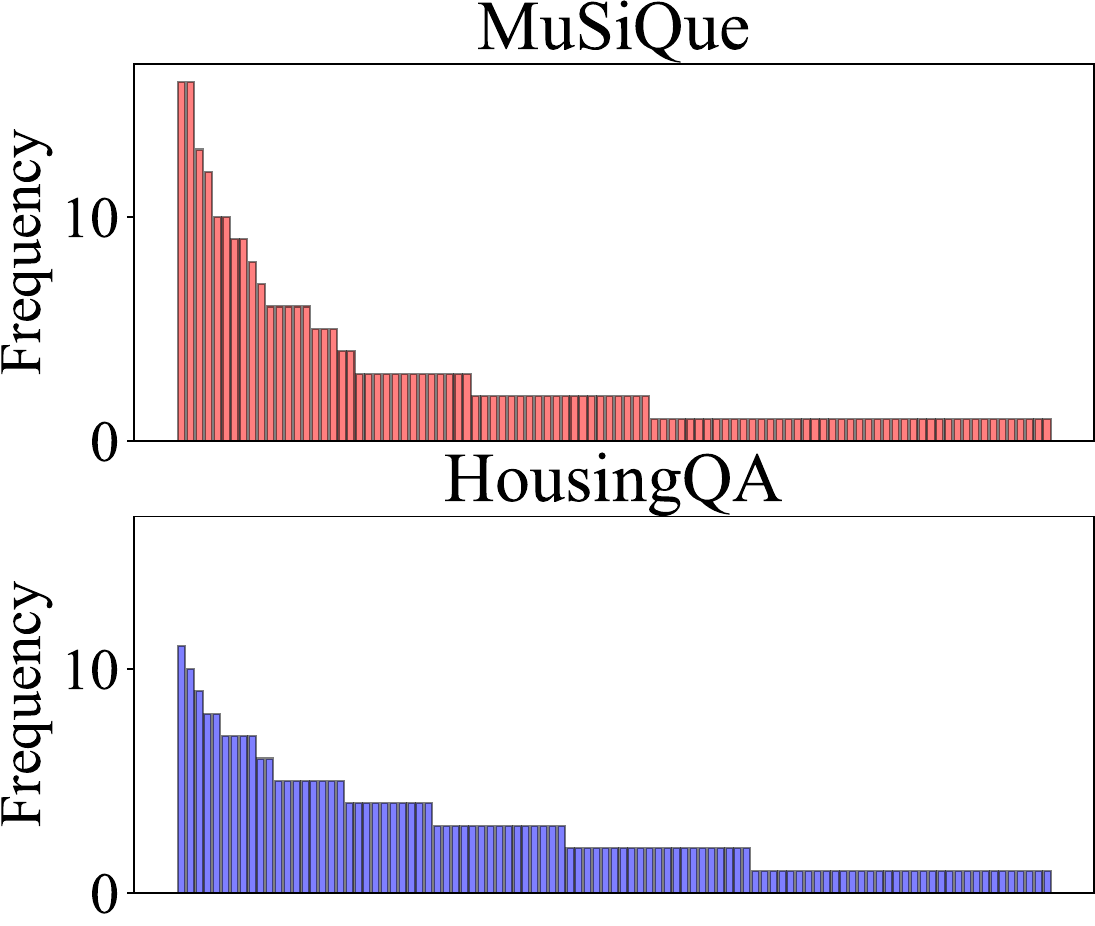}
        \vspace{-0.10in}
        \caption{\small Template frequencies.}
        \label{fig:tid_usage_histogram_musique_housingqa}
    \end{minipage}\hfill
    \begin{minipage}{0.70\linewidth}
        \centering
        \includegraphics[width=\linewidth]{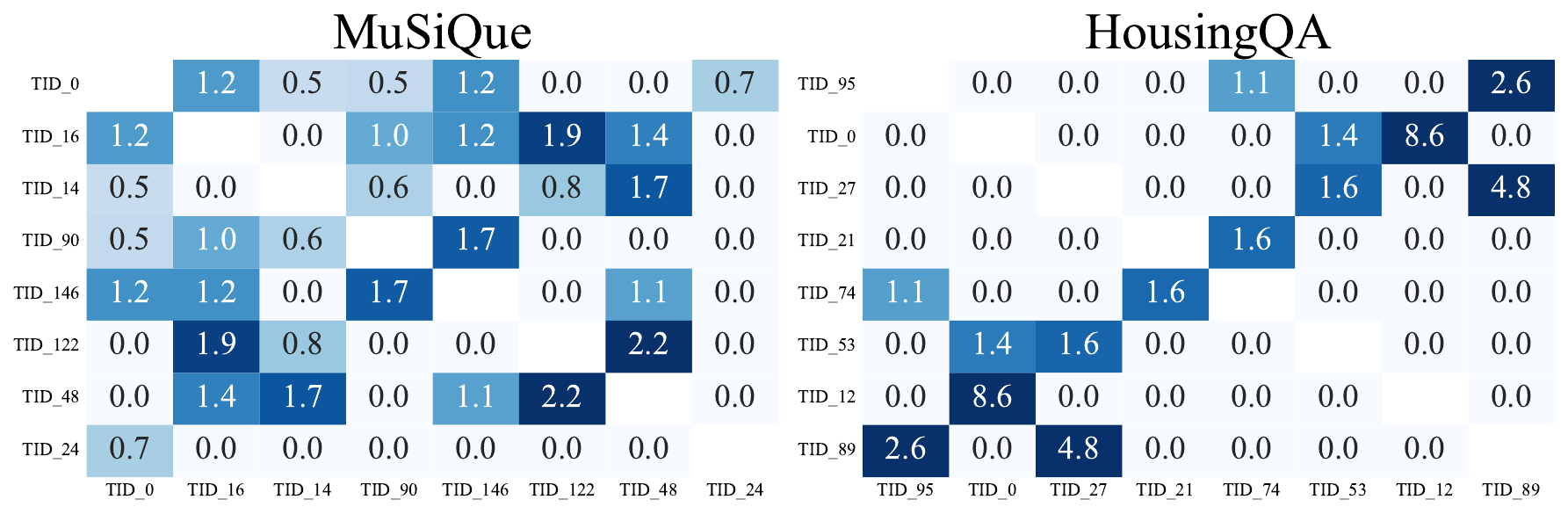}
        \vspace{-0.2in}
        \caption{\small Template co-occurrence heatmap of lift values for MuSiQue and Housing QA.}
        \label{fig:fig_template_heatmaps_two}
    \end{minipage}
    \vspace{0in}
\end{figure*}

\paragraph{Usage Distribution and Co-occurrence.}
Figure~\ref{fig:tid_usage_histogram_musique_housingqa} visualizes the distribution of template usage (Figure~\ref{fig:tid_usage_histogram_all} with all datasets).  
We observe a pronounced long-tail pattern: a small number of templates are reused frequently across many queries, while the majority are invoked only occasionally.  
This distribution reveals the coexistence of general-purpose and specialized reasoning flows.
To better understand template interactions, we compute pairwise co-occurrence statistics using lift values in Figure~\ref{fig:fig_template_heatmaps_two} (Figure~\ref{fig:fig_template_heatmaps} with all datasets), where a lift greater than 1 indicates above-chance co-usage.  
Several pair of templates exhibit consistently high lift, implying that certain reasoning templates are frequently recombined as stable compositional units, reflecting recurring reasoning routines.  
While datasets such as MuSiQue, CRAG, and FanOutQA display numerous template pairs with moderate lift values, signifying flexible and diverse reasoning combinations, Housing QA shows a contrasting trend: only a few pairs exhibit extremely high lift, with most others near independence.  
This suggests that legal domain queries are based on more rigid and repetitive reasoning structures, forming domain-specific ``template bundles'' rather than varied compositional patterns. 
Moreover, analyzing the top 10 most frequently co-occurring templates from MuSiQue reveals that 7 out of 10 originate from different training queries rather than a single source. This indicates that highly reused templates capture \textit{reusable} reasoning primitives that can be flexibly recombined across queries to handle new questions.
We also provide the example where two templates co-occurred in three different queries in Figure~\ref{fig:templates_plus_three_examples}.

\subsection{Qualitative Study}
We present a case study example in Table~\ref{tab:case_study}.  
Given the query \textit{“Why did Roncalli leave the place where Crucifixion’s creator died?”}, both \textsc{CiC} and \textsc{ToTAL} had access to the same document set, specifically documents 359 and 228.
\textsc{CiC} correctly identifies Titian as the creator and notes Roncalli’s departure from Venice but fails to connect these pieces of information, concluding that it is unanswerable.  
In contrast, \textsc{ToTAL} leverages templates to decompose the reasoning process into explicit, interpretable steps: (1) attributing the artwork to its creator, (2) locating the creator’s biographical and geographical context, and (3) linking these facts to the corresponding historical event.  
This structured reasoning chain enables the model to infer the answer correctly using the same evidence of \textsc{CiC}, highlighting how \textit{thought templates provide the missing connective reasoning that bridges retrieved facts} into a coherent multi-hop explanation.
Another example illustrating the effectiveness of the template update strategy is given in Appendix~\ref{appen:case}.

\section{Related Work}

\paragraph{Long-Context Language Models}
Along with recent advances in LLMs, there has been substantial progress in extending their input capacity, now reaching hundreds of thousands or even millions of tokens~\cite{gpt5, claude4, Gemini2.5} through architectural advances mechanisms~\cite{Su2021RoFormerET, Longformer, BigBird, Mamba, DBLP:conf/acl/0001WY025}. This enables paradigm shifts such as placing all retrieved evidence into a single prompt, scaling many-shot in-context learning to unprecedented sizes~\cite{loft, DBLP:conf/acl/BaekLGODK25, lclm_survey, DBLP:conf/acl/SeoBLH25, Byerly2024SelfConsistencyFS, DBLP:journals/corr/abs-2601-07226}, with recent benchmarks further probing these capabilities~\cite{LooGLE, DBLP:conf/acl/ZhangCHXCH0TW0024, LongBench100, LongBenchv2, loft}.

\paragraph{Reasoning with Thought Templates}
Reasoning has been a central focus in improving the capabilities of LMs. Specifically, Chain-of-Thought prompting~\cite{cot, zeroshotcot} demonstrated that explicitly eliciting intermediate reasoning steps can largely enhance model performance, and diverse variants of it have been explored~\cite{DBLP:conf/nips/YaoYZS00N23, DBLP:conf/acl/WangXLHLLL23, DBLP:conf/iclr/ZhouSHWS0SCBLC23, DBLP:conf/naacl/KongZCLQSZWD24,sot}. Building on this, recent research explores augmenting LMs with structured reasoning patterns, often referred to as thought templates. For example, \citet{BoT} proposed a framework that stores reasoning traces for math problem solving in a separate buffer and retrieves them when tackling new problems. \citet{ReasonFlux} extended it to hierarchical reasoning by identifying optimal template paths via reinforcement learning. Similarly, \citet{DBLP:journals/corr/abs-2411-18478} employed Monte Carlo Tree Search to explore reasoning trajectories, while \citet{SuperCorrect} distilled such strategies into smaller models with direct preference optimization. However, these approaches remain restricted to narrow domains (e.g., math) or rely on relatively simple reasoning steps without incorporating factual knowledge. Moreover, they typically retrieve a single template from an external buffer, whereas ours enables LCLMs to compose multiple templates simultaneously within a single generation process.

\paragraph{Text Gradient}
As directly updating or accessing the parameters of recent LMs is largely infeasible, a line of recent work has introduced natural-language feedback as a surrogate for gradients, effectively treating the LM itself as an optimizer~\cite{ProTeGi, opro, TextGrad, Cui2025ASO, DBLP:journals/corr/abs-2505-09666}. Whereas prior approaches primarily refine task or system prompts based on such feedback, our work instead updates reusable reasoning patterns by treating templates as learnable units refined through textual gradients.

\section{Conclusion}
We have presented \ours, a novel framework that fully leverages the capabilities of LCLMs by incorporating thought templates. \ours enables LCLMs to go beyond passive evidence consumption by combining factual documents with reusable reasoning patterns, and further refines these patterns through textual gradients without modifying model parameters.
\ours consistently outperforms standard prompting and RAG baselines on diverse knowledge-intensive multi-hop reasoning benchmarks, across both idealized settings without retrieval and practical scenarios with retrieval, demonstrating the effectiveness of structured reasoning guidance within long-context settings.  
Our analyses reveal that thought templates not only improve factual accuracy but also exhibit meaningful compositionality, transferability across models, and domain-awareness.  
These findings collectively highlight a promising direction for augmenting LCLMs with reusable reasoning scaffolds, transforming them from passive knowledge consumers into strategy-driven reasoners.

\section*{Limitations}
It is worth noting that our method achieves clear gains by combining compositional thought template design with iterative feedback-based refinement for complex knowledge-intensive tasks. 
Nevertheless, it assumes the availability of training queries and answers for template construction. 
In low-resource domains, this requirement may not be easily satisfied, and possible solutions include bootstrapping techniques or synthetic data generation. 
Second, within our template update framework, the feedback is produced by an auxiliary model, which can be biased or noisy, potentially leading to suboptimal refinement; thus, exploring mitigation strategies could be an interesting direction for future work. 
Finally, while our current design focuses on textual templates, extending the framework to more structured templates or multimodal contexts could further broaden its applicability.

\section*{Ethics Statement}
As our approach feeds LCLMs with a large amount of evidence documents (or sometimes the entire corpus), there is a possibility that some of these may contain harmful, sensitive, or personally identifiable information. 
We recommend that practitioners remain mindful of such risks and consider incorporating bias detection and mitigation strategies when deploying our method.

%\input{Sections/11_acknowledgement}

\bibliography{custom}

\clearpage
\appendix

\section{Experimental Results}

\begin{table}[t]
\centering\small
\caption{Iteration-wise decision summary of template updates. }
\vspace{-0.1in}
\label{tab:decision_summary}
\renewcommand{\arraystretch}{1.0}
\renewcommand{\tabcolsep}{2mm}
\resizebox{0.95\linewidth}{!}{%
\begin{tabular}{c c c c c c c}
    \toprule
    &  & \textbf{KEEP} & \textbf{ADD} & \textbf{FIX} & \textbf{DISCARD} & \textbf{F1} \\
    \midrule \midrule
    
    \multirow{5}{*}{\rotatebox{90}{\textbf{MuSiQue}}} 
    & \textbf{\textsc{CiC}}   & - & - & - & - & 63.87 \\
    & \textbf{Iter. 0}        & - & - & - & - & 70.51 \\
    & \textbf{Iter. 1}        & 4 & 0 & 10 & 0 & 71.39 \\
    & \textbf{Iter. 2}        & 2 & 1 & 9  & 0 & 73.30 \\
    & \textbf{Iter. 3}        & 1 & 0 & 7  & 0 & 71.07 \\
    
    \midrule
    
    \multirow{5}{*}{\rotatebox{90}{\textbf{CRAG}}} 
    & \textbf{\textsc{CiC}}   & - & - & - & - & 17.32 \\
    & \textbf{Iter. 0}        & - & - & - & - & 27.60 \\
    & \textbf{Iter. 1}        & 1 & 0 & 14 & 0 & 28.61 \\
    & \textbf{Iter. 2}        & 0 & 0 & 14 & 2 & 30.08 \\
    & \textbf{Iter. 3}        & 1 & 0 & 15 & 0 & 25.55 \\
    
    \bottomrule
\end{tabular}
}
\vspace{0in}
\end{table}

\subsection{Template Statistics}
We generate thought templates from 50 question–answer pairs, using a detailed prompt shown in Figure~\ref{fig:prompt_template_construction_compositional}. 
The initial template pool consists of 172 templates for MuSiQue, 162 for CRAG, 133 for FanOutQA, and 149 for HousingQA. 
These templates are then iteratively updated, as described in the following subsection.

\subsection{Template Update Strategy}
\label{appen:template_iteration_operation}
To refine the initial pool of reasoning templates, we adopt an iterative update strategy guided by textual-gradient signals. 
At each iteration, candidate templates are categorized into four actions: \texttt{KEEP}, \texttt{ADD}, \texttt{FIX}, or \texttt{DISCARD}. 

Table~\ref{tab:decision_summary} summarizes the update dynamics on MuSiQue and CRAG. 
In early iterations, the majority of updates correspond to \texttt{FIX}. 
As iterations proceed, the ratio of \texttt{KEEP} and \texttt{ADD} increases, indicating that high-quality templates become more stable and occasionally expand with new variants. 

This dynamic is also reflected in performance: on MuSiQue, F1 steadily improves up to the second iteration but slightly declines afterward, while on CRAG, moderate gains appear until the second iteration before over-refinement in the third iteration leads to a drop. 
These results suggest that the update process converges, with templates stabilizing into reusable reasoning patterns and only limited gains beyond a few iterations, resembling the diminishing returns observed in typical ML training.

\subsection{Efficiency and Cost Analysis}
We report the average token usage and number of API calls per template in Table~\ref{tab:efficiency_template}. Importantly, both template construction and update procedures are performed entirely offline and do not incur any additional cost at inference time.

While improving efficiency or latency is not the primary objective of our work, we would like to note that \textsc{ToTAL} does not introduce additional model calls at inference time, as template usage is handled within a single forward pass of the LCLM. Consequently, the per-query inference cost is comparable to \textsc{CiC}, aside from a moderate increase in input and output tokens (Table~\ref{tab:efficiency_cic}).

\begin{table}[t!]
\centering\small
\caption{Cost per template for construction and update.}
\vspace{-0.075in}
\label{tab:efficiency_template}
\renewcommand{\arraystretch}{0.9}
\renewcommand{\tabcolsep}{4mm}
\resizebox{\linewidth}{!}{%
\begin{tabular}{lcc}
    \toprule
    \textbf{Metric} & \textbf{Construction} & \textbf{Update} \\
    \midrule
    \midrule
    \textbf{Input Tokens}  & 129.76 & 1996 \\
    \textbf{Output Tokens} & 288.64 & 342 \\
    \textbf{API Calls}     & 1      & 1.57 \\
    \bottomrule
\end{tabular}
}
%\vspace{-0.025in}
\end{table}
\begin{table}[t!]
\centering\small
\caption{Cost per query for \textsc{CiC} and \textsc{ToTAL} during inference.}
\vspace{-0.075in}
\label{tab:efficiency_cic}
\renewcommand{\arraystretch}{0.9}
\renewcommand{\tabcolsep}{5mm}
\resizebox{\linewidth}{!}{%
\begin{tabular}{lcc}
    \toprule
    \textbf{Metric} & \textbf{\textsc{CiC}} & \textbf{\textsc{ToTAL}} \\
    \midrule
    \midrule
    \textbf{API Calls}     & 1      & 1 \\
    \textbf{Input Tokens}  & 122792 & 164679 \\
    \textbf{Output Tokens} & 294    & 607 \\
    \bottomrule
\end{tabular}
}
\vspace{-0.025in}
\end{table}

\subsection{Qualitative Study on Template Update}
\label{appen:case}
We further investigate the effectiveness of template update qualitatively. As shown in Figure~\ref{fig:fig_template_diff}, the original TID\_91 was decided to be \texttt{FIX} because feedback revealed that it often broke the reasoning chain in multi-hop settings, as showsn in Figure~\ref{fig:tid91_feedback}. In particular, it failed to properly integrate the output of previous steps (e.g., resolving a township into its containing county) and gave only vague instructions such as “select the relevant adjacent territory based on context,” without specifying how to apply additional constraints like naming requirements. After being updated, the revised TID\_91$'$ addressed these weaknesses by explicitly incorporating the reference territory from upstream steps, adding a filtering stage to enforce contextual conditions, and validating that the final selected territory satisfied all requirements.

Table~\ref{tab:tab_update_casestudy} illustrates the effect of this refinement. With the original TID\_91, the model incorrectly answered “Oconto Falls,” as it failed to maintain continuity with the birthplace entity and did not filter adjacent candidates. In contrast, the updated TID\_91$'$ produced the correct answer “Green Bay,” demonstrating that the refined template successfully chained outputs across steps and enforced query-specific constraints. This case highlights that our template update strategy makes the reasoning process more consistent and reliable in multi-hop queries.

\begin{table*}[t]
\centering
\caption{\small Case study on the multi-hop query from MuSiQue.}
\label{tab:case_study}
\vspace{-0.1in}
\renewcommand{\arraystretch}{1.1}
\setlength{\tabcolsep}{6pt}
{\small
\begin{tabular}{p{0.15\linewidth} p{0.75\linewidth}}
\toprule
\textbf{Query} & Why did Roncalli leave the place where Crucifixion's creator died? \\
\midrule
\textbf{Gold Answer} & for the conclave in Rome \\
\midrule
\textbf{\textsc{CiC}} & 
Looking at the documents to trace this query:\newline

TITLE: Crucifixion (Titian) | ID: 359\newline
TITLE: Pope John XXIII | ID: 228\newline

From document ID: 359, I can see that the Crucifixion was created by Titian, a Venetian artist, and the painting is located in Ancona. However, this document doesn't specify where Titian died.\newline

From document ID: 228, I can see that Roncalli (who became Pope John XXIII) left Venice for the conclave in Rome, and it mentions he had a return train ticket to Venice, suggesting he expected to return but didn't because he was elected Pope.\newline

However, the documents don't explicitly state where Titian died or provide a clear connection between Titian's death location and Roncalli's departure from that place. The query cannot be fully answered with the information provided in these documents.\newline

\textcolor{red}{Final Answer: [`Cannot be determined from the provided documents']} \\
\midrule
\textbf{\textsc{ToTAL} (Ours)} & 
Looking at this query, I need to identify who created "Crucifixion," where they died, and why Roncalli left that place.\newline

Step 1 | TEMPLATE\_TITLE: Work-to-Creator Attribution TEMPLATE\_ID: TID\_77 | TEMPLATE\_CONTENT: Descriptions: Finding the creator or contributor associated with a specific creative work such as films, books, or other media. Reason Flow: [`Identify the specific creative work in question', `Determine what type of contributor information is needed (screenwriter, director, etc.)', `Look up the creative credits or attribution for that work', "Extract the specific person's name who fulfilled that role"] Example: {`example\_problem': "Who was the screenwriter of 'With the Mounted Police'?", 'solution\_steps': ["Identify the film: `With the Mounted Police'", `Look up the screenwriting credits for this film', `Find that Lloyd Lonergan was the screenwriter'], `final\_answer': `Lloyd Lonergan'}\newline

Step 2 | TEMPLATE\_TITLE: Biographical Location Lookup TEMPLATE\_ID: TID\_58 | TEMPLATE\_CONTENT: Descriptions: Finding the specific location associated with a significant life event of a person, such as where they died, were born, or lived. Reason Flow: [`Identify the specific person in question', `Determine what type of location information is needed (death, birth, residence, etc.)', `Research biographical records or databases', `Extract the specific location associated with that life event'] Example: {`example\_problem': `In what city did Lloyd Lonergan die?', `solution\_steps': [`Identify the person: Lloyd Lonergan', `Research biographical information about his death', `Find records indicating he died in New York'], `final\_answer': `New York'}\newline

Step 3 | TEMPLATE\_TITLE: Historical Event Specification TEMPLATE\_ID: TID\_139 | TEMPLATE\_CONTENT: Descriptions: Determining specific historical events or circumstances related to a known figure, particularly focusing on significant life events. Reason Flow: [`Focus on the specific aspect being questioned (e.g., cause of death, major events)', `Research the historical record for that specific detail', `Identify the commonly accepted historical fact'] Example: {`example\_problem': `What was the cause of death of Jesus?', `solution\_steps': [`Focus on the specific question: cause of death', `Research historical accounts of Jesus's death', `Identify the commonly accepted cause: crucifixion'], `final\_answer': `crucifixion'}\newline

TITLE: Crucifixion (Titian) | ID: 359\newline
TITLE: Pope John XXIII | ID: 228\newline

\textcolor{green!60!black}{Final Answer: [`for the conclave in Rome']} \\
\bottomrule
\end{tabular}
}
\end{table*}

\clearpage
\appendix

% --- TID_91 Before/After ---
\begin{figure*}[t]
\centering

% ===================== TID_91 Original =====================
\begin{minipage}[t]{0.48\textwidth}
\begin{tcolorbox}[colback=gray!5, colframe=gray!50,
                  boxrule=0.5pt, arc=2mm,
                  left=4pt, right=4pt, top=3pt, bottom=3pt]
\footnotesize
\textbf{TID\_91 — Adjacent Territory Identification (Original)}\\[2pt]
\emph{Description.} Finding administrative territories that share borders or are adjacent to a given territory.\\[4pt]
\emph{Reason flow.}
\begin{enumerate}[topsep=0pt, itemsep=0ex, parsep=0pt, left=12pt]
  \item Identify the reference territory
  \item Search for border or adjacency relationships
  \item List all territories that share borders
  \item Select the relevant adjacent territory based on context
\end{enumerate}
\emph{Example.}
\begin{itemize}[topsep=0pt, itemsep=0.2ex, parsep=0pt, left=12pt]
  \item \textbf{Problem:} Which county shares a border with Lincoln County?
  \item \textbf{Solution steps:}
  \begin{itemize}[topsep=0pt, itemsep=0ex, parsep=0pt, left=12pt]
    \item Identify the reference territory: Lincoln County
    \item Search for counties that share borders with Lincoln County
    \item Identify Nye County as one that shares a border with Lincoln County
  \end{itemize}
  \item \textbf{Final answer:} \texttt{Nye County}
\end{itemize}
\end{tcolorbox}
\end{minipage}
\hfill
% ===================== TID_91 Revised =====================
\begin{minipage}[t]{0.48\textwidth}
\begin{tcolorbox}[colback=gray!5, colframe=gray!50,
                  boxrule=0.5pt, arc=2mm,
                  left=4pt, right=4pt, top=3pt, bottom=3pt]
\footnotesize
\textbf{TID\_91$'$ — Adjacent Territory Identification (Revised)}\\[2pt]
\emph{Description.} Finding administrative territories that share borders or are adjacent to a given territory, including cases where the reference territory must first be determined from contained entities.\\[4pt]
\emph{Reason flow.}
\begin{enumerate}[topsep=0pt, itemsep=0ex, parsep=0pt, left=12pt]
  \item Identify or receive the reference territory from previous steps
  \item If reference territory contains sub-entities, confirm the containing territory
  \item Search for all territories that share borders with the reference territory
  \item Apply additional filtering criteria from the query context
  \item Validate that selected adjacent territory meets all constraints
  \item Select the final adjacent territory that matches all requirements
\end{enumerate}
\emph{Example.}
\begin{itemize}[topsep=0pt, itemsep=0.2ex, parsep=0pt, left=12pt]
  \item \textbf{Problem:} Which county shares a border with Dearborn County and is named after a river?
  \item \textbf{Solution steps:}
  \begin{itemize}[topsep=0pt, itemsep=0ex, parsep=0pt, left=12pt]
    \item Identify the reference territory: Dearborn County
    \item Search for all counties that share borders with Dearborn County
    \item List adjacent counties: Ohio, Ripley, Franklin
    \item Apply filtering: must be named after a river
    \item Validate: Ohio County is named after Ohio River
    \item Select Ohio County
  \end{itemize}
  \item \textbf{Final answer:} \texttt{Ohio County}
\end{itemize}
\end{tcolorbox}
\end{minipage}

\medskip
\caption{\small Comparison of \textbf{TID\_91} before and after refinement.}
\label{fig:fig_template_diff}
\end{figure*}

% --- TID_91 Feedback ---
\begin{figure*}[t]
\centering
\begin{tcolorbox}[colback=gray!3, colframe=gray!50,
                  boxrule=0.5pt, arc=2mm,
                  left=6pt, right=6pt, top=4pt, bottom=4pt, width=0.9\textwidth, breakable]
\footnotesize
\textbf{Feedback for TID\_91 (from textual-gradient signals).}\\[4pt]

\begin{itemize}[topsep=0pt, itemsep=0.4ex, parsep=0pt, left=12pt]
  \item Failed to chain with previous step (Miller Township $\rightarrow$ Dearborn County).
  \item Jumped to irrelevant result (\emph{Río de la Plata}) instead of adjacent counties.
  \item Missing handling of multi-step adjacency (contained entities).
  \item Vague filtering step: \emph{“Select based on context”}.
  \item Needs explicit integration with prior outputs.
  \item Missing validation against query constraints (e.g., named after a river).
\end{itemize}

\end{tcolorbox}
\caption{\small Textual-gradient feedback guiding the refinement of \textbf{TID\_91}.}
\label{fig:tid91_feedback}
\end{figure*}

\begin{table*}[t]
\centering
\caption{\small Case study on showing the effect of refining \textbf{TID\_91}.}
\label{tab:tab_update_casestudy}
\vspace{-0.1in}
\renewcommand{\arraystretch}{1.1}
{\small
\begin{tabular}{p{0.15\linewidth} p{0.75\linewidth}}
\toprule
\textbf{Query} & What city is near the birthplace of Erik Jensen (American football) and shares a border with that place? \\
\midrule
\textbf{Gold Answer} & Green Bay \\
\midrule
\textbf{Before (TID\_91)} & 
Step 1 | TEMPLATE\_TITLE: Biographical Location Lookup \; TEMPLATE\_ID: TID\_48 \; TEMPLATE\_CONTENT: Descriptions: Finding specific geographic locations associated with important life events of a person ... \newline

Step 2 | TEMPLATE\_TITLE: Administrative Capital Lookup \; TEMPLATE\_ID: TID\_145 \; TEMPLATE\_CONTENT: Descriptions: Finding the capital city or administrative center of a given political entity ... \newline

Step 3 | TEMPLATE\_TITLE: Adjacent Territory Identification \; TEMPLATE\_ID: \textbf{TID\_91} \; TEMPLATE\_CONTENT: Descriptions: Finding administrative territories that share borders or are adjacent to a given territory ... \newline

Step 4 | TEMPLATE\_TITLE: Administrative Capital Lookup \; TEMPLATE\_ID: TID\_145 \; TEMPLATE\_CONTENT: Descriptions: ... \newline

TITLE: Erik Jensen (American football) | ID: 367 \newline
TITLE: Pulaski High School | ID: 26 \newline
TITLE: Kelly Brook, Wisconsin | ID: 371 \newline

\textcolor{red}{Final Answer: [`Oconto Falls']} \\
\midrule
\textbf{After (TID\_91$'$)} & 
Step 1 | TEMPLATE\_TITLE: Person-to-Birthplace Mapping \; TEMPLATE\_ID: TID\_105 \; TEMPLATE\_CONTENT: Descriptions: Finding the birthplace or place of origin of a specific person ... \newline

Step 2 | TEMPLATE\_TITLE: Adjacent Territory Identification \; TEMPLATE\_ID: \textbf{TID\_91$'$ (Revised)} \; TEMPLATE\_CONTENT: Descriptions: Includes confirmation of containing territories and applying filters ... \newline

Step 3 | TEMPLATE\_TITLE: Administrative Capital Lookup \; TEMPLATE\_ID: TID\_145 \; TEMPLATE\_CONTENT: Descriptions: Finding the capital city or administrative center of a given political entity ... \newline

TITLE: Erik Jensen (American football) | ID: 367 \newline
TITLE: Pulaski High School | ID: 26 \newline

\textcolor{green!60!black}{Final Answer: [`Green Bay']} \\
\bottomrule
\end{tabular}
}
\end{table*}

\begin{figure*}
    \centering
    \includegraphics[width=0.975\linewidth]{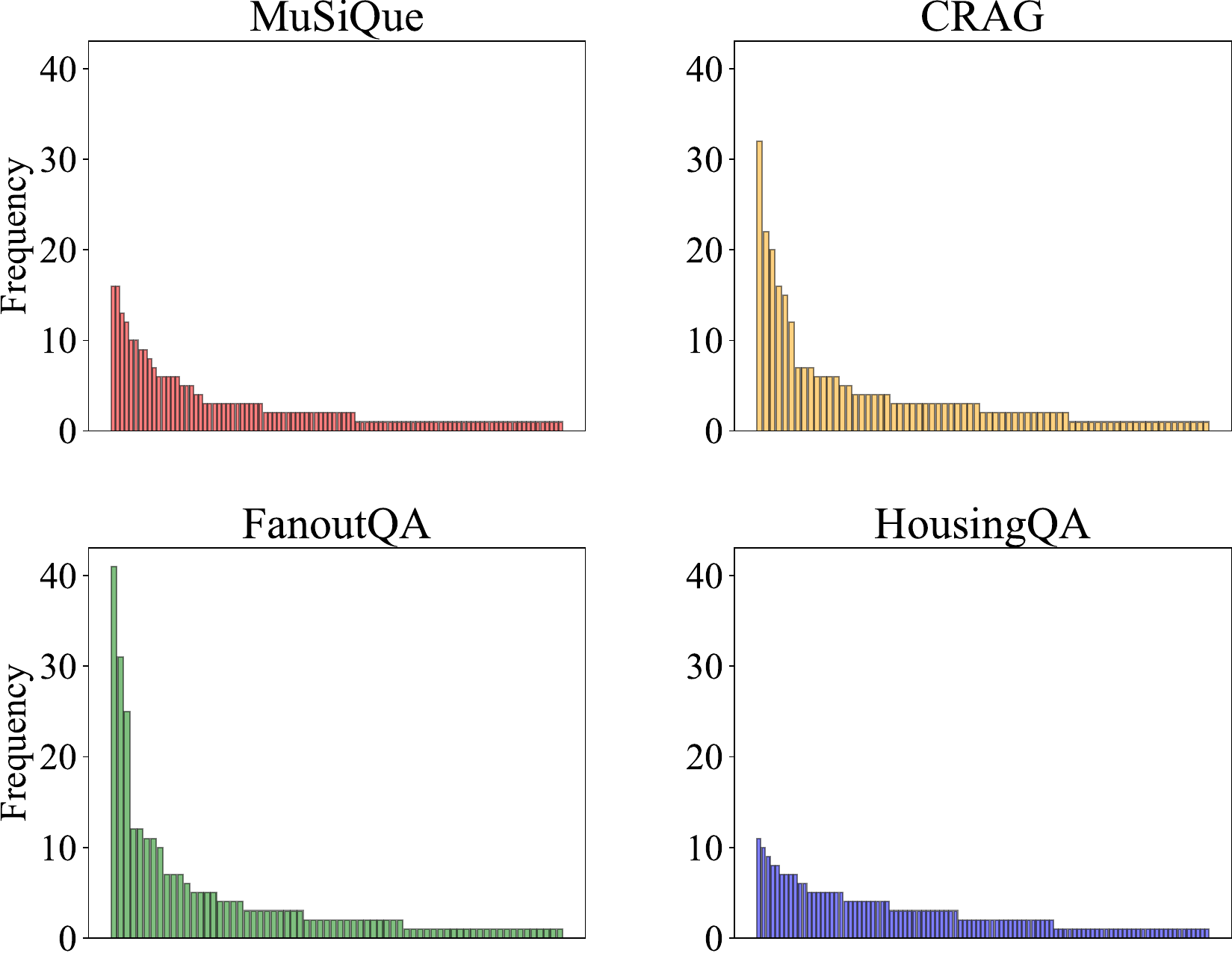}
    \vspace{-0.075in}
    \caption{\small Histogram of template frequencies across datasets.
    }
    \label{fig:tid_usage_histogram_all}
    \vspace{-0.1in}
\end{figure*}
\begin{figure*}
    \centering
    \includegraphics[width=0.98\linewidth]{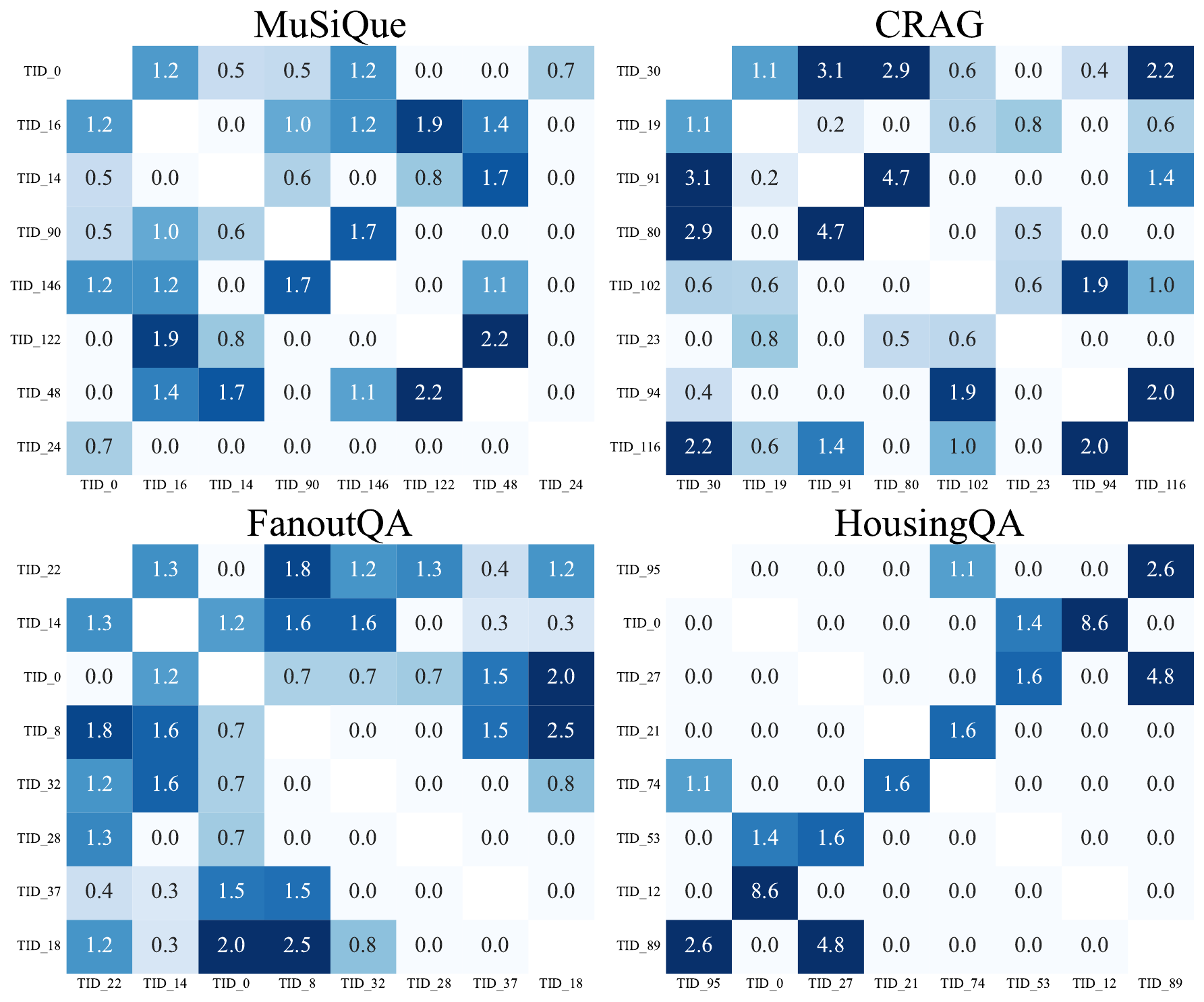}
    \vspace{-0.075in}
    \caption{\small Template co-occurrence heatmap of lift values across datasets.
    }
    \label{fig:fig_template_heatmaps}
    \vspace{-0.1in}
\end{figure*}

\begin{figure*}[t]
\centering

% ===================== Templates (side-by-side) =====================
\begin{minipage}[t]{0.48\textwidth}
\begin{tcolorbox}[colback=gray!5, colframe=gray!50,
  boxrule=0.5pt, arc=2mm, left=4pt, right=4pt, top=3pt, bottom=3pt]
\footnotesize
\textbf{TID\_45 — Administrative Territory Identification}\\[2pt]
\textbf{Description.} A method for determining which administrative territorial entity (state, province, country) contains a specific location.\\[4pt]
\textbf{Reason flow.}
\begin{enumerate}[topsep=0pt, itemsep=0ex, parsep=0pt, left=12pt]
  \item Identify the specific location
  \item Determine the type of administrative division needed (state, province, country, etc.)
  \item Identify which administrative entity contains that location
\end{enumerate}
\textbf{Example.}
\begin{itemize}[topsep=0pt, itemsep=0.2ex, parsep=0pt, left=12pt]
  \item \textbf{Problem:} What state is Boston located in?
  \item \textbf{Solution steps:}
  \begin{itemize}[topsep=0pt, itemsep=0ex, parsep=0pt, left=12pt]
    \item Identify the location: Boston
    \item Determine the administrative level needed: state level
    \item Identify the containing administrative entity: Massachusetts
  \end{itemize}
  \item \textbf{Final answer:} \texttt{Massachusetts}
\end{itemize}
\end{tcolorbox}
\end{minipage}
\hfill
\begin{minipage}[t]{0.48\textwidth}
\begin{tcolorbox}[colback=gray!5, colframe=gray!50,
  boxrule=0.5pt, arc=2mm, left=4pt, right=4pt, top=3pt, bottom=3pt]
\footnotesize
\textbf{TID\_65 — Demographic Ranking and Selection}\\[2pt]
\textbf{Description.} Finding the top-ranked entity within a geographic region based on specific demographic criteria.\\[4pt]
\textbf{Reason flow.}
\begin{enumerate}[topsep=0pt, itemsep=0ex, parsep=0pt, left=12pt]
  \item Identify the geographic scope for comparison
  \item Determine the ranking criteria (population, area, etc.)
  \item Apply the criteria to find the top-ranked entity
  \item Verify the result meets all specified conditions
\end{enumerate}
\textbf{Example.}
\begin{itemize}[topsep=0pt, itemsep=0.2ex, parsep=0pt, left=12pt]
  \item \textbf{Problem:} What city is Russia's largest metropolitan area as measured by population?
  \item \textbf{Solution steps:}
  \begin{itemize}[topsep=0pt, itemsep=0ex, parsep=0pt, left=12pt]
    \item Identify that we need the largest metropolitan area in Russia
    \item Apply population-based ranking criteria
    \item Determine that Moscow has the largest metropolitan population in Russia
  \end{itemize}
  \item \textbf{Final answer:} \texttt{Moscow}
\end{itemize}
\end{tcolorbox}
\end{minipage}

\medskip

% ===================== All three queries in ONE box =====================
\begin{tcolorbox}[colback=gray!5, colframe=gray!50,
  boxrule=0.5pt, arc=2mm, left=4pt, right=4pt, top=3pt, bottom=3pt, width=\textwidth]
\footnotesize
\textbf{Queries using these templates (chains only).}

\medskip
\textbf{Query 1.} Who won the Indy Car Race in the largest populated city of the state where the performer of \emph{Mingus Three} is from?\\[-2pt]
\textbf{Templates applied:} TID\_14 $\rightarrow$ TID\_105 $\rightarrow$ TID\_45 $\rightarrow$ TID\_65 \quad
\textbf{Final Answer:} \texttt{Mario Andretti}

\medskip
\textbf{Query 2.} What was the wettest year in the second largest city in the state where Yuma's Library District is located?\\[-2pt]
\textbf{Templates applied:} TID\_45 $\rightarrow$ TID\_65 $\rightarrow$ TID\_90 \quad
\textbf{Final Answer:} \texttt{1905}

\medskip
\textbf{Query 3.} How long are the city council terms in the second largest city in the state where Yuma is located?\\[-2pt]
\textbf{Templates applied:} TID\_45 $\rightarrow$ TID\_65 $\rightarrow$ TID\_66 \quad
\textbf{Final Answer:} \texttt{four-year terms}
\end{tcolorbox}

\caption{Two templates (\textbf{TID\_45}, \textbf{TID\_65}) and three queries that used these templates simultaneously.}
\label{fig:templates_plus_three_examples}
\end{figure*}

\begin{figure*}[t]
\centering
\begin{tcolorbox}[colback=gray!5, colframe=gray!50,
                  boxrule=0.5pt, arc=2mm,
                  left=4pt, right=4pt, top=3pt, bottom=3pt]
\textbf{Template Construction Prompt.}

You are an expert in reasoning strategies. Given a complex, multi-step problem, its complete solution, and the final answer, extract a structured problem-solving template composed of reusable sub-templates. Return the result in JSON format with the following structure:

\begin{enumerate}[topsep=0pt, itemsep=0ex, parsep=0pt, left=14pt]
  \item A clear name for the strategy (\texttt{template\_name})
  \item A brief description of the method (\texttt{description})
  \item A step-by-step reasoning flow to solve similar problems (\texttt{reason\_flow})
  \item An example application, including:
  \begin{itemize}[topsep=0pt, itemsep=0ex, parsep=0pt, left=12pt]
     \item Problem statement (\texttt{example\_problem})
     \item Solution steps (\texttt{solution\_steps})
     \item Final answer (\texttt{final\_answer})
  \end{itemize}
  \item \texttt{sub\_templates}: A list of dictionaries, each representing a reasoning sub-template with:
  \begin{itemize}[topsep=0pt, itemsep=0ex, parsep=0pt, left=12pt]
     \item \texttt{template\_name}: A descriptive name for this sub-strategy
     \item \texttt{description}: A brief description of the sub-strategy
     \item \texttt{reason\_flow}: A list of reasoning steps involved in this sub-task
     \item \texttt{example}: An example application of this sub-template, including:
     \begin{itemize}[topsep=0pt, itemsep=0ex, parsep=0pt, left=12pt]
        \item \texttt{example\_problem}: A question matching this reasoning pattern
        \item \texttt{solution\_steps}: Step-by-step solution to that question
        \item \texttt{final\_answer}: The answer to that question
     \end{itemize}
  \end{itemize}
\end{enumerate}

\medskip
\textbf{Instruction constraint:} Respond only in JSON format with no explanation.

\medskip
\textbf{Inputs shown to the model:}
\begin{verbatim}
Problem:
""" {problem} """

Solution:
""" {solution} """

Final Answer:
""" {answer} """
\end{verbatim}
\end{tcolorbox}
\caption{Prompt used to construct compositional thought templates from (Problem, Solution, Final Answer). Each generated sub-template is treated as a template and added to $\mathcal{T}$.
}
\label{fig:prompt_template_construction_compositional}
\end{figure*}

\begin{figure*}[t]
    \centering
    \begin{tcolorbox}[colback=gray!5, colframe=gray!50,
                      boxrule=0.5pt, arc=2mm, 
                      left=4pt, right=4pt, top=3pt, bottom=3pt]
\textbf{Role.} You are improving a reasoning template where it was applied.

\textbf{Current Template.}
\begin{verbatim}
{JSON dump of the current template:
 "template_id","template_name","description","reason_flow","example"}
\end{verbatim}

\textbf{Failed Cases where this template was used.}
\begin{verbatim}
Case #0 (F1: {f1})
Query: {query text}
REASONING TRACE: {model_outputs[0]}
Gold: {gold answer}
Pred: {prediction}
\end{verbatim}

\textbf{Failed Case Source (original query/solution/answer).}
\begin{verbatim}
Query: {problem}
Solution Steps:
{step-by-step solution or evidence block}
Final Answer: {answer}
\end{verbatim}

\textbf{Your task.} Analyze the template's role in the prediction error:
\begin{itemize}[topsep=0pt, itemsep=0ex, parsep=0pt, left=10pt]
  \item How the template led to the incorrect prediction
  \item What needs to be fixed in the template
  \item Specific feedback to get the correct answer
\end{itemize}

\textbf{Decision Guide (choose exactly one at the end).}
\begin{itemize}[topsep=0pt, itemsep=0ex, parsep=0pt, left=10pt]
  \item \textbf{FIX} – Template needs revision to address the issues above
  \item \textbf{DISCARD} – Template is fundamentally incorrect
  \item \textbf{KEEP} – Template works perfectly AND failure is due to external factors (e.g., answer format)
  \item \textbf{ADD} – Template works perfectly BUT failure is due to system coordination issues (e.g., selection, multi-step integration)
\end{itemize}

\textbf{Output format.}
\begin{itemize}[topsep=0pt, itemsep=0ex, parsep=0pt, left=10pt]
  \item Return bullets only for your analysis.
  \item On the \textbf{FINAL LINE}, output \textbf{exactly one} of: \texttt{**FIX**} or \texttt{**DISCARD**} or \texttt{**ADD**} or \texttt{**KEEP**}.
\end{itemize}
    \end{tcolorbox}
    \caption{Prompt for generating textual gradient feedback.}
    \label{fig:prompt_textgrad_feedback}
\end{figure*}

\begin{figure*}[t]
    \centering
    \begin{tcolorbox}[colback=gray!5, colframe=gray!50,
                      boxrule=0.5pt, arc=2mm,
                      left=4pt, right=4pt, top=3pt, bottom=3pt]
\textbf{Role.} You will edit a reasoning template based on the FEEDBACK.

\textbf{Output constraints.}
\begin{itemize}[topsep=0pt, itemsep=0ex, parsep=0pt, left=10pt]
  \item Return \textbf{ONLY} a valid JSON object matching the SCHEMA below.
  \item \textbf{No markdown, no extra text.} Use double quotes for all keys/strings.
\end{itemize}

\textbf{SCHEMA.}
\begin{verbatim}
{
  "template_id": "string",
  "template_name": "string",
  "description": "string",
  "reason_flow": ["string", "..."],
  "example": {
    "example_problem": "string",
    "solution_steps": ["string", "..."],
    "final_answer": "string"
  }
}
\end{verbatim}

\textbf{Current Template.}
\begin{verbatim}
{JSON dump of the current template:
 "template_id","template_name","description","reason_flow","example"}
\end{verbatim}

\textbf{Failed Cases (referenced in feedback).}
\begin{verbatim}
Case #0 (F1: {f1})
Query: {query text}
REASONING TRACE: {model_outputs[0]}
Gold: {gold answer}
Pred: {prediction}
\end{verbatim}

\textbf{Failed Case Source (original query/solution/answer).}
\begin{verbatim}
Query: {problem}
Solution Steps:
{step-by-step solution or evidence block}
Final Answer: {answer}
\end{verbatim}

\textbf{FEEDBACK.} (from Fig.~\ref{fig:prompt_textgrad_feedback})
\begin{verbatim}
{feedback text}
\end{verbatim}

\textbf{Instruction.} Revise the template to address the FEEDBACK while preserving reusable structure and staying within the SCHEMA. \textbf{Respond only with the JSON object.}
    \end{tcolorbox}
    \caption{Prompt for template update given textual gradient feedback.}
    \label{fig:prompt_textgrad_edit}
\end{figure*}

\end{document}